\def\DROFullVersion{1}
\documentclass[twocolumn,amsthm]{autart}

\usepackage{ifthen}
\newboolean{showcomments}
\setboolean{showcomments}{true}
\usepackage[usenames,dvipsnames]{color}

\definecolor{bleudefrance}{rgb}{0.19, 0.55, 0.91}
\definecolor{ao(english)}{rgb}{0.0, 0.5, 0.0}

\newcommand{\addcite}[0]{\ifthenelse{\boolean{showcomments}}
{\textcolor{purple}{(add cite(s)) }}{}}%

\newcommand{\you}[1]{  \ifthenelse{\boolean{showcomments}}
{{\color{blue} PCY: #1}}{}}
\newcommand{\youmargin}[1]{\ifthenelse{\boolean{showcomments}}{\marginpar{\color{bleudefrance}\tiny PCY: #1}}{}}

\newboolean{showedits}
\setboolean{showedits}{true}
\usepackage[markup=underlined]{changes}
\definechangesauthor[color=bleudefrance]{PCY}
\newcommand{\magentatext}[1]{%
\ifthenelse{\boolean{showedits}}
{\textcolor{magenta}{#1}}
{#1}}
\newcommand{\ayou}[1]{
\ifthenelse{\boolean{showedits}}
{\added[id=PCY]{#1}}
{\!#1\hspace{-4.75pt}}
}
\newcommand{\ryou}[2]{
\ifthenelse{\boolean{showedits}}
{\replaced[id=PCY]{#1}{#2}}
{\!#1\hspace{-4.75pt}}
}
\newcommand{\dyou}[1]{
\ifthenelse{\boolean{showedits}}
{\deleted[id=PCY]{#1}}
{}
}
\newcommand{\hl}[1]{\ifthenelse{\boolean{showcomments}}
{\textcolor{purple}{#1}}{}}%

\usepackage[labelformat=simple]{subcaption}

\graphicspath{{figures/}}

\newcommand{\xdro}{x^{\mathtt{DRO}}}
\newcommand{\Jdro}{J^{\mathtt{DRO}}}
\newcommand{\Josp}{J^{\mathtt{OSP}}}
\newcommand{\empP}[1]{\widehat{\bbP}_{#1}^{N_{#1}}}

\input{edits_yzliu.tex}

\ifdefined\DROFullVersion
  \newcommand{\proofdeferred}[1]{%
    \par\noindent\textit{Proof.} See Appendix~\ref{#1}.\par
  }

  \newcommand{\mref}[1]{\ref{#1}}
  \newcommand{\meqref}[1]{\eqref{#1}}
  \makeatletter
  \def\ps@copyright{\let\@mkboth\@gobbletwo
    \def\@oddhead{}\let\@evenhead\@oddhead
    \def\@oddfoot{\small\slshape Preprint\hfil\thepage}
    \let\@evenfoot\@oddfoot}
  \makeatother
\else
  \usepackage{xr}
  \newcommand{\proofdeferred}[1]{%
    \par\noindent\textit{Proof.} See Appendix~\ref{apx-#1}.\par
  }

\fi

\graphicspath{{figures/}}

\renewcommand{\thepage}{\arabic{page}}
\begin{document}

\begin{frontmatter}

\runtitle{Distributionally Robust Federated Averaging with Multi-Source Data}

\title{Communication-Efficient Distributionally Robust Federated Averaging with Multi-Source Data}

\thanks[funding]{This work was supported in part by the National Science and Technology Major Project under grant 2022ZD0116401, and in part by the National Natural Science Foundation of China under grants 72431001, U2241214, and
62373008. AC was supported in part by TKI HTSM FD-CODE (24PPS175).}
\author[pku]{Yingzhu Liu},
\ead{yzliucoe@pku.edu.cn}
\author[pku]{Zhongkui Li},
\ead{zhongkli@pku.edu.cn}
\author[pku]{Pengcheng You\corauthref{cor}},
\ead{pcyou@pku.edu.cn}
\corauth[cor]{Corresponding author.}
\author[ug]{Ashish Cherukuri}
\ead{a.k.cherukuri@rug.nl}

\address[pku]{Department of Control Science and Systems Engineering, Peking University}
\address[ug]{Engineering and Technology Institute Groningen, University of Groningen}


\begin{abstract}
Federated learning trains a shared model from distributed client data. 
The deployment risk is typically specified through two approximations: each unknown local distribution is replaced by its empirical distribution, and the unknown deployment mixture across clients is replaced by empirical or prescribed training weights. 
Existing robust methods usually address one approximation at a time through local distributional ambiguity sets or worst-case client mixtures. 
However, joint treatment of both approximations remains underexplored. 
To address this gap, we construct a global ambiguity set formed by admissible mixtures of client-wise Wasserstein ambiguity sets, yielding a distributionally robust optimization formulation. 
The construction allows client-dependent ambiguity radii, preserves client-wise separability of the objective function, and yields a high-probability out-of-sample guarantee for the resulting robust problem.
We then introduce a penalty-based approximation of this formulation and develop a communication-efficient algorithm called BiDRO-FedAvg, for which we establish a convergence guarantee. We validate the proposed framework through a market-demand prediction simulation.
\end{abstract}

\end{frontmatter}

\let\AND\relax

\section{Introduction}
\begin{table*}[t]
\centering
\caption{High-level comparison of ambiguity modeling and federated implementation.}
\label{tab:intro-method-comparison}
\scriptsize
\setlength{\tabcolsep}{3.5pt}
\renewcommand{\arraystretch}{1.08}

\begin{tabularx}{\textwidth}{
    @{}
    >{\raggedright\arraybackslash}p{0.27\textwidth}
    Y Y Y Y Y
    @{}
}
\toprule
&
\multicolumn{3}{c}{\textbf{Ambiguity modeling}}
&
\multicolumn{2}{c}{\textbf{Federated implementation}}
\\
\cmidrule(lr){2-4}
\cmidrule(l){5-6}

\textbf{Method}
&
\makecell{\textbf{Local}\\\textbf{ambiguity}}
&
\makecell{\textbf{Mixture}\\\textbf{uncertainty}}
&
\makecell{\textbf{Client-dependent}\\\textbf{radius}}
&
\makecell{\textbf{Partial client}\\\textbf{participation}}
&
\makecell{\textbf{Multiple}\\\textbf{local steps}}
\\
\midrule

FedAvg~\cite{mcmahan2017communication}
&
$\times$
&
$\times$
&
$\times$
&
$\checkmark$
&
$\checkmark$
\\
\addlinespace[0.4mm]

AFL~\cite{mohri2019agnostica}
&
$\times$
&
$\checkmark$
&
$\times$
&
$\times$
&
$\times$
\\
\addlinespace[0.4mm]

DRFA~\cite{deng2020distributionally}
&
$\times$
&
$\checkmark$
&
$\times$
&
$\checkmark$
&
$\checkmark$
\\
\addlinespace[0.4mm]

DRO-FL~\cite{wang2025distributionally}
&
$\checkmark$
&
$\checkmark$
&
$\times$
&
$\times$
&
$\times$
\\
\addlinespace[0.4mm]

Group DRO with distributional uncertainty~\cite{konti2025group}
&
$\checkmark$
&
$\checkmark$
&
$\checkmark$
&
--
&
--
\\
\addlinespace[0.4mm]

\textbf{BiDRO-FedAvg (ours)}
&
$\checkmark$
&
$\checkmark$
&
$\checkmark$
&
$\checkmark$
&
$\checkmark$
\\

\bottomrule
\end{tabularx}

\vspace{0.5mm}
\parbox{\textwidth}{%
\scriptsize
\emph{Note:}
``--'' denotes that the feature is not applicable to the centralized setting.
}
\end{table*}
Federated learning (FL) enables a central server to train a shared model across multiple clients without collecting their raw data~\cite{mcmahan2017communication}. This paradigm is well suited to applications in which data are naturally distributed and privacy-sensitive, such as mobile services and healthcare networks~\cite{antunes2022federated}. In practice, the deployment risk can be specified through two empirical approximations, which induce two sources of statistical uncertainty. First, limited local samples may fail to accurately represent the true local distributions, leading to within-client distributional ambiguity. Second, when clients have heterogeneous local distributions, the unknown deployment mixture becomes a critical source of group-level uncertainty. In addition, communication is often limited, and not all clients may participate in every communication round. These challenges hinder the development of robust and well-generalized models, and also call for communication-efficient algorithms.

To handle within-client distributional ambiguity, distributionally robust optimization (DRO) models uncertainty through an ambiguity set around the empirical distribution{~\cite{mohajerinesfahani2018datadriven,kuhn2019wasserstein,cherukuri2020cooperative,blanchet2019quantifying}}.In particular, Wasserstein ambiguity sets are commonly used since they capture geometric perturbations and often admit tractable reformulations{~\cite{mohajerinesfahani2018datadriven,gao2023distributionally,shafieezadehabadeh2019regularization}}.
In multi-client settings, some related approaches  build a single global ambiguity set around a representative distribution, such as a Wasserstein barycenter or a specified mixture of local distributions{~\cite{lau2022wasserstein,nguyen2022generalization}}. Although this yields a compact global description, local uncertainties are absorbed into one shared ambiguity-set construction. As a result, the relation between the global ambiguity set and the underlying local distributions becomes less transparent.
It also becomes harder to choose a single radius that is large enough to cover plausible deployment distributions, yet small enough to avoid excessive conservatism. An alternative is to construct separate Wasserstein balls around the local empirical distributions and form an ambiguity set from mixtures of distributions within these balls~\cite{ibrahim2025fdrsvm}, which preserves client-wise distributional ambiguity. However, this formulation uses fixed mixture weights and therefore does not account for uncertainty in the deployment mixture, which is central in federated learning.

Complementary to modeling local ambiguity, another challenge is how to define a coherent global learning objective for multiple clients. Classical FL algorithms, such as FedAvg~\cite{mcmahan2017communication}, use empirical sample proportions or prescribed mixture weights for aggregation. However, a mismatch between the training and deployment mixtures can misalign the objective with deployment risk and bias the learned model. To address this issue, several approaches optimize over worst-case mixtures{~\cite{zecchin2022communicationefficient}}. For instance, agnostic federated learning (AFL) improves robustness and fairness by considering the worst-case mixture of local objectives~\cite{mohri2019agnostica}, while distributionally robust federated averaging (DRFA) adopts a similar worst-mixture perspective~\cite{deng2020distributionally}. Beyond the FL setting, Group DRO similarly optimizes over the worst-case mixture of local distributions under group-level shifts~\cite{soma2022near,yu2024efficient}. However, all these methods robustify the mixture weights while still evaluating local losses on empirical client distributions. As a result, they do not explicitly account for within-client distributional ambiguity, which can be significant when local distributions are estimated from finite samples.

Recently, several works have begun to jointly consider within-client ambiguity and mixture uncertainty. In federated learning, an outlier-robust formulation based on a specially designed unbalanced Wasserstein metric addresses both uncertainty levels~\cite{wang2025distributionally}. However, it uses a single global ambiguity radius and requires full client participation at every communication round. In a related multi-source DRO setting, a group distributionally robust formulation incorporates both uncertainty levels and allows source-specific ambiguity radii, but its formulation and solution algorithm are centralized~\cite{konti2025group}. As a result, extending it to federated training with partial participation and reduced communication remains nontrivial. Together, these limitations show that adopting jointly robust formulations to practical federated training remains challenging.
Partial client participation and infrequent communication further complicate the optimization.

In this paper, we develop a distributionally robust FL framework based on a global ambiguity set formed by admissible mixtures of client-wise Wasserstein ambiguity sets.
The construction provides a common basis for the out-of-sample analysis and our proposed federated algorithm BiDRO-FedAvg.
Table~\ref{tab:intro-method-comparison} compares the ambiguity-modeling and federated-implementation features of representative methods and BiDRO-FedAvg.
We further validate the proposed algorithm through a market-demand prediction study under both shifts in local distributions and client mixtures.
Our main contributions are summarized as follows:
\begin{enumerate}
    \item We construct a global ambiguity set as admissible mixtures of client-wise Wasserstein ambiguity sets. The construction robustifies both empirical local distributions and the client mixture, allows client-dependent local radii, and admits a client-wise separable reformulation.
    \item We establish high-probability coverage and out-of-sample performance guarantees for the resulting problem, characterizing the effects of local sample sizes, local ambiguity radii, client heterogeneity, and mixture-weight deviations.
    \item We introduce a penalized formulation with an induced-radius robustness interpretation, and we develop an algorithm BiDRO-FedAvg with partial client participation and multiple local updates. We establish an \(\mathcal{O}(T^{-3/8}+\epsilon)\) bound on its expected primal-dual gap.
\end{enumerate}

A preliminary conference version of this work is to appear as~\citet{liu2026distributionally}. The present paper substantially extends it in theory, algorithm design, and numerical validation.
We establish a robust interpretation of the penalized local problems. For each fixed decision, a worst-case distribution of each penalized problem is also worst-case over a Wasserstein ball with an induced radius. The algorithm in~\citet{liu2026distributionally}, which performs one local model update per client per communication round, is redesigned as BiDRO-FedAvg to support partial participation and multiple local updates, together with a new convergence analysis. The numerical study is also substantially expanded to a market-demand prediction problem with simultaneous shifts in local distributions and client mixtures.
The evaluation includes multiple communication protocols, sensitivity analyses, and empirical convergence verification.%
\ifdefined\DROSubmissionLongVersionNote\space\DROSubmissionLongVersionNote\fi

\textbf{Notations:}
Let $\bbR$ and $\bbR_{\geq0}$ denote the set of real and nonnegative real numbers.
Let $\bbN_+:=\{1,2,\ldots\}$ denote the set of positive integers.
For \(n\in\mathbb N_+\), write $[n]:=\{1,2,\ldots,n\}$ and let $\Delta_{n-1}:=\{\alpha\in\bbR^n:~\alpha_i\geq0,\sum_{i=1}^n\alpha_i=1\}$ be the $n-1$ probability simplex.
Let $\|\cdot\|$ denote the Euclidean 2-norm.
The $i$-th component of a vector $v\in\bbR^n$ is denoted by $v_i$.
For a finite set $A$, $\mathrm{Unif}(A)$ denotes the uniform distribution over $A$.
\section{Problem statement}
Consider a federated system with $n$ clients indexed by $[n]$.
Each client updates its local model using private data and communicates only with a central server to train a shared model.
Let $\bbP$ denote the \emph{unknown} true group distribution of clients. 
Let $\xi$ be a measurable random variable with the support $\Xi\subseteq\bbR^m$ that follows the distribution $\bbP$, where the set $\Xi$ is convex and compact.
The true distribution $\bbP$ is assumed to be a mixture of local distributions:
\begin{align}\label{eq:mixture4ground_truth}
    \bbP=\sum_{i=1}^n \alpha_i^\star \bbP_i\,,
\end{align}
where $\alpha^\star:=(\alpha^\star_1,\ldots,\alpha_n^\star)\in\Delta_{n-1}$ is the \emph{unknown} mixture weights, and $\bbP_i$ is the true local distribution supported on $\Xi$ for client $i\in [n]$.
Client $i$ gathers a local dataset $\calD_i:=\{\widehat{\xi}_i^k\}_{k=1}^{N_i}$ consisting of $N_i$ samples  independently drawn from the \emph{unknown} local distribution $\bbP_i$.
The local datasets $\{\mathcal D_i\}_{i=1}^n$ are mutually independent across clients.
Note that $\alpha^\star_i$ may differ from the empirical proportions $N_i/N$, where $N:=\sum_iN_i$ is the total number of samples.
The goal is to learn a shared model \(x\in\calX\) from distributed data, where \(\calX\subseteq\bbR^d\) is a nonempty, convex, and compact feasible set with diameter
\(D_x:=\max_{u,v\in\calX}\|u-v\|\). Let $\ell:\bbR^d\times\Xi\rightarrow\bbR$ be a loss function that is continuous on \(\calX\times\Xi\), where \(\ell(x,\xi)\) denotes the loss incurred by model \(x\) on data sample \(\xi\).
The performance of the learned model is evaluated under the true group distribution $\bbP$, i.e., $\ev_{\bbP}[\ell(x,\xi)]$.
For example, in federated supervised learning, \(\xi=(u,y)\) represents a feature--label pair, \(x\) the parameters of a predictive model or a neural network, and \(\ell(x,\xi)\) a standard loss such as the squared error or cross-entropy.

\subsection{Within-client distributional ambiguity}
Given the observed samples, each client can approximate $\bbP_i$ by the empirical distribution $\empP{i}$ from local samples, given by
\begin{align}\label{eq:empirical_distri}
    \empP{i}:=\frac{1}{N_i}\sum_{k=1}^{N_i}\delta_{\widehat{\xi}_i^k}\,,
\end{align}
where $\delta_{\widehat{\xi}_i^k}$ is the unit point mass at $\widehat{\xi}_i^k$.
However, such an estimator may not accurately represent $\bbP_i$ with limited data. 
Additionally, the single point estimate may fail to capture local perturbations around the true distribution $\bbP_i$.
To address both issues, we construct a set of plausible distributions for each client $i$, referred to as an ambiguity set, and optimize against the worst-case distribution in this set.
Specifically, for a fixed $p\in[1,\infty)$ and each client $i\in [n]$, we define the local ambiguity set as a $p$-Wasserstein ball of radius $\varepsilon_i$ centered at $\empP{i}$:
\begin{align}\label{eq:local_ambiguity_set}
    \calW_i(\varepsilon_i):=\Bigl\{\bbQ\in\calM(\Xi):W_p(\bbQ,\empP{i})\leq\varepsilon_i\Bigr\}\,,
\end{align}
where $\calM(\Xi)$ is the set of probability distributions on $\Xi$ with finite $p$-th moment.
Moreover, $\varepsilon_i\ge 0$ is a client-dependent ambiguity radius, reflecting different levels of local uncertainty across clients. 
    Here, $W_p$ denotes the $p$-Wasserstein distance on the Polish metric space $(\Xi,d)$. The metric $d$ is assumed to induce the relative Euclidean topology on $\Xi$.
    For probability distributions $\bbP,\bbP^\prime\in\calM(\Xi)$ , the distance is defined as~\cite{kuhn2019wasserstein}
\begin{align}\label{eq:def_wp}
W_p(\bbP,\bbP^\prime):=
\left(
\inf_{\gamma\in\Gamma(\bbP,\bbP^\prime)}
\int_{\Xi\times\Xi} c(x,y)\,\mathrm d\gamma(x,y)
\right)^{1/p},
\end{align}
where $\Gamma(\bbP,\bbP^\prime)$ is the set of all couplings with marginals $\bbP$ and $\bbP^\prime$, and $c(x,y):=d(x,y)^p$ is the transportation cost.

\subsection{Cross-client mixture uncertainty}
\begin{figure}[htbp]
    \centering
    \includegraphics[width=0.75\columnwidth]{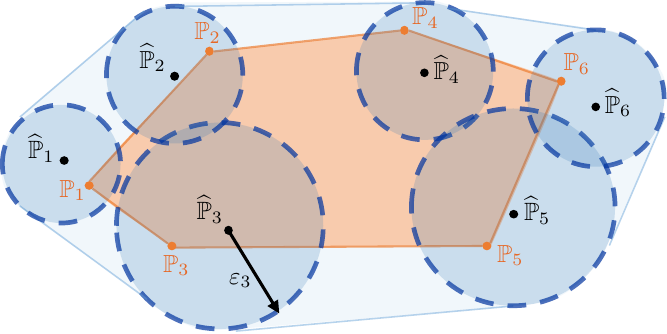}
    \caption{Illustration of the ambiguity set $\calW_{\mathrm G}$ defined in~\eqref{eq:global_ambiguity_set} for $n=6$ clients. For client $i$, the orange dot denotes the true distribution $\bbP_i$, the black dot denotes the associated empirical distribution $\empP{i}$, and the dashed blue ball centered at $\widehat{\bbP}_i$ represents the local Wasserstein ambiguity set $\calW_i(\varepsilon_i)$ with radius $\varepsilon_i$.
    The light-blue region represents $\calW_{\mathrm G}$, obtained by mixing candidate distributions selected from the local ambiguity sets over all admissible client weights. The orange-shaded polygon represents all mixtures of true local distributions.
    }
    \label{fig:ambiguity_set}
\end{figure}

The true group distribution in~\eqref{eq:mixture4ground_truth} is obtained by aggregating the local distributions according to the mixture weights~$\alpha^\star$. Hence, for a fixed $\alpha\in\Delta_{n-1}$, a natural group-level ambiguity model preserves the mixture structure while accounting for client-level uncertainty.
Specifically, allowing each client's distribution to vary within its Wasserstein ambiguity set leads to the following group-level ambiguity set:
\begin{align}\label{eq:global_ambiguity}
    \calW_{\mathrm M}(\alpha):=\Big\{\sum_{i=1}^n\alpha_i\mathbb{Q}_i:\ \mathbb{Q}_i\in\calW_i(\varepsilon_i),\forall i\in[n]\Big\}\,,
\end{align}
where $\alpha_i$ is the weight associated with client $i$. Equivalently, $\calW_{\mathrm M}(\alpha)=\sum_{i=1}^{n}\alpha_i\calW_i(\varepsilon_i)$, which is the weighted Minkowski sum of the local ambiguity sets.
Ideally, if the true weights $\alpha^\star$  were known, one would use $\calW_{\mathrm M}(\alpha^\star)$ to construct the global ambiguity set. 
However, since $\alpha^\star$ is unavailable in practice, a natural robust approach is to consider all mixtures induced by weights in the feasible set $\Delta_{n-1}$. Accordingly, we define the global ambiguity set as a union:
\begin{align}
        &\calW_{\mathrm G}
:=\bigcup\nolimits_{\alpha\in\Delta_{n-1}}\calW_{\mathrm M}(\alpha)
\notag\\
&=\Big\{\sum_{i=1}^n\alpha_i\bbQ_i:\ \alpha\in\Delta_{n-1},\ \bbQ_i\in\calW_i(\varepsilon_i),\forall i\in[n]\Big\}.\label{eq:global_ambiguity_set}
\end{align}
This construction jointly captures within-client distributional ambiguity through $\calW_i(\varepsilon_i)$ and cross-client mixture ambiguity through the union over $\alpha\in\Delta_{n-1}$.
Fig.~\ref{fig:ambiguity_set} visualizes the geometry of the proposed global ambiguity set for $n=6$ clients.
The blue regions enclosed by the dashed circles depict the local ambiguity sets while the orange-shaded polygon represents the set of all mixtures of the true local distributions.
As illustrated in Fig.~\ref{fig:ambiguity_set}, if $\bbP_i\in\calW_i(\varepsilon_i)$ for every client, then all mixtures of the true local distributions, including $\bbP$, belong to the global ambiguity set $\calW_{\mathrm G}$.
\subsection{Problem formulation}
With the global ambiguity set $\calW_{\mathrm G}$ in place, we formulate the following minimax problem, referred to as the mixture Wasserstein distributionally robust optimization problem:
\begin{align}\tag{MW-DRO}\label{eq:problem_DRO}
    \min_{x\in\calX}\max\limits_{\bbQ\in\calW_{\mathrm G}} \ev_{\bbQ}[\ell(x,\xi)]\,,
\end{align}
where $x\in\calX\subseteq\bbR^d$ is the decision variable. 
The goal is to seek a solution that is robust to both within-client distributional ambiguity and uncertainty in the client mixture weights.
Let $\xdro$ denote an optimal solution of~\eqref{eq:problem_DRO}, and let $\Jdro$ be its optimal value.
We further define the out-of-sample performance (OSP) of the solution $\xdro$ as $\Josp:=\ev_{\bbP}[\ell(\xdro,\xi)]$, namely, its expected loss under the true distribution $\bbP$.
This leads to two central questions:
\begin{enumerate}
    \item \textbf{Out-of-sample guarantee:} 
    How large is the performance gap between the optimal value $\Jdro$ of~\eqref{eq:problem_DRO} and the true expected loss $\Josp$ (both evaluated at the solution $\xdro$)?
    \item \textbf{Efficient computation:} Can we solve the problem~\eqref{eq:problem_DRO} efficiently in a federated setting with partial client participation and limited communication?
\end{enumerate}
\section{Out-of-sample performance guarantees}
In this section, since the solution of~\eqref{eq:problem_DRO} is ultimately deployed under the true distribution $\bbP$, we examine whether the optimal value $\Jdro$ provides a reliable certificate for the out-of-sample performance $\Josp$.
First, we establish a finite-sample \emph{coverage} guarantee ensuring~$\bbP$ belongs to the proposed global ambiguity set with high probability. Then the coverage result ensures that $\Josp\leq\Jdro$ with high probability and yields an explicit upper bound on $\Jdro-\Josp$.
\subsection{Coverage of the global ambiguity set}
To establish coverage of the global ambiguity set, we first control the probability that each true local distribution belongs to its corresponding local ambiguity set. 
We impose the following local coverage condition.

\begingroup
\begin{assumption}[Local Empirical Coverage]\label{ass:local_coverage}
For each client $i\in[n]$ with $N_i\geq1$, there exists a known nonincreasing function $\beta_i^{N_i}:\bbR_{\geq0}\to[0,1]$ such that, for every $\varepsilon\geq0$,
\begin{align}
\bbP_i^{\otimes N_i}\!
\left(
W_p(\bbP_i,\empP{i})
\le \varepsilon
\right)
\ge
1-\beta_i^{N_i}(\varepsilon).
\label{eq:local_empirical_coverage}
\end{align}
\end{assumption}
\endgroup
{The following remark gives one such bound for the Euclidean metric.}
\begin{remark}\label{rem:local_coverage}
For the Euclidean ground metric $d(\xi,\zeta)=\|\xi-\zeta\|_2$, let $R:=\frac{1}{2}\operatorname{diam}_{\infty}(\Xi)$. If $R>0$, one such bound is~\cite{fournier2015rate,boskos2024highconfidence}
\begin{equation}\label{eq:local_concentration_bound}
\beta_i^{N_i}(\varepsilon)
:=\min\left\{1,\,
c_1\exp\left[-c_2N_i\,
\psi_{p,m}\left(\frac{\varepsilon}{R}\right)\right]\right\},
\end{equation}
where $c_1\geq1$ and $c_2>0$ depend only on $p$ and $m$, and
\[
\psi_{p,m}(u):=
\begin{cases}
{h(u^p)}, & p=m/2,\\[1mm]
u^{\max\{2p,m\}}, & p\neq m/2,
\end{cases}
\]
where $h(u):=u^2/(\ln(2+1/u))^2$ for $u>0$, with $\psi_{p,m}(0):=0$. For $m>2p$, explicit choices of the constants are discussed in~\citet[Remark~4.3]{boskos2024highconfidence}. If $R=0$, then $\Xi$ is a singleton, and one can take $\beta_i^{N_i}(\varepsilon)=0$ for all $\varepsilon\geq0$. \oprocend
\end{remark}
By Assumption~\ref{ass:local_coverage}, for a given local radius $\varepsilon_i$, the true local distribution \(\bbP_i\) is contained in the induced local ambiguity set with probability at least \(1-\beta_i^{N_i}(\varepsilon_i)\).
A direct route to global coverage is to require the local coverage events to hold simultaneously across all clients. However, the intersection of all local events may render the condition conservative. When the local distributions are sufficiently close, coverage of an ambiguity set centered at one client may extend to other clients. This suggests the probability of deriving a less conservative coverage bound by exploiting cross-client structure.
We formalize this idea through the following bounded-heterogeneity assumption.
\begin{assumption}~\label{ass:bounded_heterogeneity}
    Suppose there exists a constant $\delta\ge 0$ such that for all $i,j\in[n]$, $
W_p(\bbP_i,\bbP_j)\le \delta$.
\end{assumption}
With small heterogeneity $\delta$, one local ambiguity set may cover all true local distributions, yielding a tighter bound.
\begin{proposition}[Coverage of the Global Ambiguity Set]~\label{prop:coverage}
Let Assumptions~\ref{ass:local_coverage} and~\ref{ass:bounded_heterogeneity} hold. For given radii $\varepsilon_i\geq0$, $i\in[n]$, let $\beta_i:=\beta_i^{N_i}(\varepsilon_i)$ be the corresponding bounds on the local failure probabilities. Define the bounds on the local failure probabilities at heterogeneity-adjusted radii:
\[
\bar\beta_i:=
\begin{cases}
\beta_i^{N_i}(\varepsilon_i-\delta), & \varepsilon_i\ge\delta,\\1, & \varepsilon_i<\delta.
\end{cases}
\]
Then, we have
\begin{align}\label{eq:global_coverage}
    \Pr\!\left(\bbP\in\calW_{\mathrm G}\right)\geq 1-\beta,
\end{align}
where $1-\beta:=\max\left\{\prod_{i=1}^n(1-\beta_i),\quad1-\prod_{i=1}^n\bar\beta_i\right\}.$
\end{proposition}
\proofdeferred{proof:coverage}
\subsection{Performance evaluation under true group distribution}
We now connect the coverage result to the out-of-sample performance of the solution of~\eqref{eq:problem_DRO}.
The coverage implies that on the event $\{\bbP\in\calW_{\mathrm G}\}$, the worst-case expected loss in~\eqref{eq:problem_DRO} upper-bounds the true expected loss for any ${x\in\calX}$:
\begin{align}\label{eq:upper_worst}
    \ev_{\bbP}[\ell(x,\xi)] \;\le\; \sup_{\bbQ\in\calW_{\mathrm G}}\ev_{\bbQ}[\ell(x,\xi)]\,.
\end{align}
By Proposition~\ref{prop:coverage}, the optimal value $\Jdro$ provides, with probability at least $1-\beta$, a certificate for the true expected loss $\Josp$ at the solution $\xdro$.
To assess the tightness of this certificate, we next bound the out-of-sample performance gap $\Jdro-\Josp$.
A key step is to exploit the union-and-mixture structure of $\calW_{\mathrm G}$, which yields a client-wise decomposition of the objective. 
The following result formalizes the separability and yields an equivalent client-wise reformulation of~\eqref{eq:problem_DRO}.

\begin{lemma}[Problem Separability]\label{lm:separability}
The problem~\eqref{eq:problem_DRO} admits the following reformulation:
\begin{align}
    &\min_{x\in\calX} \max_{\mathbb{Q} \in \calW_{\mathrm G}} \mathbb{E}_{\mathbb{Q}} \left[ \ell(x,\xi) \right]\nonumber
\\
=&~ \min_{x\in\calX}\max_{\alpha\in\Delta_{n-1}} \sum_{i=1}^n \alpha_i
\max_{\bbQ_i\in\calW_i(\varepsilon_i)}
\mathbb{E}_{\mathbb{Q}_i} \left[ \ell(x,\xi) \right]. \label{eq:problem_separability}
\end{align}
\end{lemma}
\proofdeferred{proof:separability}
With the separable representation in place, we next bound the out-of-sample performance gap $\Jdro-\Josp$.
\begin{theorem}[Out-of-Sample Performance under Separated Coverage Events]\label{thm:OOS_separated}
Let Assumptions~\ref{ass:local_coverage} and~\ref{ass:bounded_heterogeneity} hold.
Assume that, for all $x\in\calX$, the function $\xi\mapsto \ell(x,\xi)$ is $L_{\xi}^{\ell}$-Lipschitz with respect to $d$.
Under the same setting as Proposition~\ref{prop:coverage}, with probability at least $1-\beta$,
    \begin{equation}\label{eq:oos_upper_bound}
        \Josp\leq \Jdro .
    \end{equation}
    Furthermore, on the simultaneous local-coverage event $\bigcap_{i=1}^n\left\{\bbP_i\in\calW_i(\varepsilon_i)\right\}$, which occurs with probability at least $\prod_{i=1}^n\left(1-\beta_i^{N_i}(\varepsilon_i)\right)$, we have
    \[
        \Jdro-L_\xi^\ell(2r+\delta)
        \leq \Josp\leq \Jdro,
        \qquad
        r:=\max_{i\in[n]}\varepsilon_i.
    \]
\end{theorem}
\proofdeferred{proof:oos}
\begin{remark}
Theorem~\ref{thm:OOS_separated} bounds the gap between $\Jdro$ and $\Josp$ by $L_\xi^\ell(2r+\delta)$ when all local coverage events hold simultaneously.
The two terms $r$ and $\delta$ reflect the contributions of local ambiguity radii and client heterogeneity, respectively.
In the Euclidean case,~\eqref{eq:local_concentration_bound} shows that larger local sample sizes allow smaller ambiguity radii at a fixed target coverage confidence, thereby reducing the first term and tightening the bound.
However, additional samples do not reduce the heterogeneity term when the underlying client distributions are fixed.
\oprocend
\end{remark}
\section{Penalty-based reformulation and\\ algorithm design}\label{sec:algorithm}
We now turn to solving~\eqref{eq:problem_DRO} in a federated manner. Although Lemma~\ref{lm:separability} decomposes the problem across clients, each local objective still involves an optimization over probability measures subject to a hard Wasserstein ball constraint and cannot be directly handled by standard federated stochastic gradient methods. We therefore construct a penalized formulation and derive an equivalent finite-dimensional reformulation for distributed learning.
\subsection{Tractable reformulation}
By Lemma~\ref{lm:separability}, client $i$ has the constrained local objective $\max_{\mathbb Q_i\in\calW_i(\varepsilon_i)} \ev_{\mathbb Q_i}[\ell(x,\xi)]$ in~\eqref{eq:problem_separability}.
We replace the hard constraint with a client-dependent penalty and define
\begin{align}\label{eq:local_penalized_objective}
F_i(x):=\sup_{\mathbb Q_i\in\calM(\Xi)}
\left\{
\ev_{\mathbb Q_i}[\ell(x,\xi)]-\rho_i W_p^p(\mathbb Q_i,\empP{i})
\right\}\,,
\end{align}
where the parameter $\rho_i>0$ determines the cost of deviating from $\empP{i}$ and controls the trade-off between local robustness and conservatism. A larger $\rho_i$ penalizes distributional deviations more heavily and induces a smaller effective ambiguity region by encouraging the adversarial distribution to remain closer to $\empP{i}$.
Moreover, the penalized local objective~\eqref{eq:local_penalized_objective} retains an exact DRO interpretation at an induced radius for each client $i${, as established next.}
\begin{proposition}[Robust Interpretation of the Penalized Local Problem]\label{prop:local_robustness}
For $i\in[n]$ and $\rho_i>0$, let
\begin{equation*}
    \bbP_i^{\star}(x) \mathrel{{\in}} \operatorname*{arg\,max}_{\bbQ_i\in\calM(\Xi)} \ev_{\bbQ_i}\Big[\ell(x,\xi)-\rho_i W_p^p(\bbQ_i,\empP{i}) \Big]\,,
\end{equation*}
and define the induced radius as $\widehat{r}_i(x) := W_p(\bbP_i^{\star}(x),\empP{i})$.
Then, we have
\begin{align}
    \sup_{\bbQ_i:
W_p(\bbQ_i,\empP{i})
\le
\widehat r_i(x)}
\ev_{\bbQ_i}[\ell(x,\xi)]
=
\ev_{\bbP_i^\star(x)}[\ell(x,\xi)]\,.
\end{align}
\end{proposition}
\proofdeferred{proof:local_robustness}
For each fixed \(x\), the penalty \(\rho_i\) implicitly induces a radius \(\widehat r_i(x)\) such that the worst-case distribution \(\bbP_{i}^\star(x)\) solves both the penalized local problem and the constrained DRO problem centered at \(\empP{i}\) with the induced radius \(\widehat r_i(x)\).
Thus, the penalized formulation preserves distributional robustness with respect to the induced local ambiguity balls.

Replacing each local objective in~\eqref{eq:problem_separability} by~\eqref{eq:local_penalized_objective} yields the penalized problem
\begin{align}\label{eq:penalized_DRO}
\min_{x\in\calX}\max_{\alpha\in\Delta_{n-1}}
\sum_{i=1}^n\alpha_iF_i(x).
\end{align}
However, each local objective $F_i$ still involves optimization over probability measures. 
We next derive an equivalent empirical representation of $F_i$ that can be evaluated from local samples.
Define $\phi_i(x,\zeta,\xi):=\ell(x,\xi)-\rho_i c(\zeta,\xi)$.
Using the definition in~\eqref{eq:def_wp} and the coupling formulation 
of the Wasserstein distance,  we obtain
\begin{align}
F_i(x)
&=\sup_{\bbQ_i\in\calM(\Xi)}\Big\{
\int_\Xi \ell(x,\xi)\,\mathrm d\bbQ_i(\xi)\notag\\
&\qquad-
\rho_i
\inf_{\gamma_i\in\Gamma(\empP{i},\bbQ_i)}
\int_{\Xi\times\Xi}
c(\zeta,\xi)\,
\mathrm d\gamma_i(\zeta,\xi)
\Big\}
\notag\\
&\stackrel{\mathrm{(a)}}{=}
\sup_{\substack{
\bbQ_i\in\calM(\Xi),\\
\gamma_i\in\Gamma(\empP{i},\bbQ_i)}}
\int_{\Xi\times\Xi}
\phi_i(x,\zeta,\xi)\,
\mathrm d\gamma_i(\zeta,\xi)
\notag\\
&\stackrel{\mathrm{(b)}}{=}
\sup_{\substack{
\gamma_i\in\calM(\Xi\times\Xi):\\
(\gamma_i)_\zeta=\empP{i}}}
\int_{\Xi\times\Xi}
\phi_i(x,\zeta,\xi)\,
\mathrm d\gamma_i(\zeta,\xi)
\notag\\
&\stackrel{\mathrm{(c)}}{=}
\sup_{\nu_{i,1},\ldots,\nu_{i,N_i}\in\calM(\Xi)}
\frac{1}{N_i}
\sum_{k=1}^{N_i}
\int_\Xi
\phi_i(x,\widehat\xi_i^k,\xi)\,
\mathrm d\nu_{i,k}(\xi)
\notag\\
&\stackrel{\mathrm{(d)}}{=}
\frac{1}{N_i}
\sum_{k=1}^{N_i}
\sup_{\nu\in\calM(\Xi)}
\int_\Xi
\phi_i(x,\widehat\xi_i^k,\xi)\,
\mathrm d\nu(\xi)
\notag\\
&\stackrel{\mathrm{(e)}}{=}
\frac{1}{N_i}
\sum_{k=1}^{N_i}
\sup_{\xi\in\Xi}
\phi_i(x,\widehat\xi_i^k,\xi)
\notag\\
&=
\ev_{\zeta\sim\empP{i}}
\left[f_i(x,\zeta)\right],
\label{eq:local_pointwise_reformulation}
\end{align}
where $
f_i(x,\zeta):=\sup_{\xi\in\Xi}\phi_i(x,\zeta,\xi)$.
Equality~$\mathrm{(a)}$ follows since $\bbQ_i$ is the $\xi$-marginal of every $\gamma_i\in\Gamma(\empP{i},\bbQ_i)$ and $\rho_i>0$. 
Equality~$\mathrm{(b)}$ follows since every coupling with $\zeta$-marginal $\empP{i}$ determines a second marginal $\bbQ_i=(\gamma_i)_\xi$, and every feasible pair $(\bbQ_i,\gamma_i)$ in the preceding line defines such a coupling.
For equality~$\mathrm{(c)}$, by~\eqref{eq:empirical_distri}, every coupling with $\zeta$-marginal $\empP{i}$ can be represented as
$
\gamma_i(\mathrm d\zeta,\mathrm d\xi)
=
\frac{1}{N_i}
\sum_{k=1}^{N_i}
\delta_{\widehat\xi_i^k}(\mathrm d\zeta)
\nu_{i,k}(\mathrm d\xi)$,
where $\nu_{i,k}\in\calM(\Xi)$ is the conditional distribution associated with $\widehat\xi_i^k$. Conversely, every collection $\{\nu_{i,k}\}_{k=1}^{N_i}$ defines a feasible coupling with $\zeta$-marginal $\empP{i}$.
Equality~$\mathrm{(d)}$ follows because the measures $\nu_{i,k}$ can be optimized independently. Equality~$\mathrm{(e)}$ uses
$
\sup_{\nu\in\calM(\Xi)}
\int_\Xi h(\xi)\,\mathrm d\nu(\xi)
=
\sup_{\xi\in\Xi}h(\xi)$,
since every expectation is bounded above by the pointwise supremum, while Dirac measures at maximizing points give the reverse inequality. The final equality follows from
the definitions of $f_i$ and $\empP{i}$.

Substituting~\eqref{eq:local_pointwise_reformulation} into~\eqref{eq:penalized_DRO} for every $i\in[n]$ yields the equivalent reformulation of~\eqref{eq:penalized_DRO}
\begin{align}\label{eq:reformed_DRO}
\min_{x\in\calX}\max_{\alpha\in\Delta_{n-1}}F(x,\alpha),
\end{align}
where $F(x,\alpha)
:=\sum_{i=1}^n\alpha_i
\ev_{\zeta\sim\empP{i}}[f_i(x,\zeta)]$.
The objective has a natural distributed structure as a weighted sum of client-specific empirical objectives.
\subsection{Algorithm design}
To solve the problem~\eqref{eq:reformed_DRO} under partial client participation and limited communication, we present BiDRO-FedAvg, a distributionally robust federated averaging algorithm. Each communication round of BiDRO-FedAvg consists of model broadcast, client selection, and local model updates at the participating clients, model aggregation, and a mixture-weight update.

We next introduce the notation used to state the algorithm.
Let $\tau\in\bbN_+$ denote the communication interval, defined as the number of local model-update iterations per communication round.
Over $S\in\bbN_+$ communication rounds, the total number of model-update iterations is $T:=S\tau$.
We index these iterations consecutively by $t=0,\ldots,T-1$.
A communication round may include multiple message exchanges between the server and clients.
For each round \(s\in\{0,\ldots,S-1\}\), define the iteration block as $\mathcal I_s:=\{s\tau,\ldots,(s+1)\tau-1\}$.
Let $s_t:=\lfloor t/\tau\rfloor$ denote the communication round containing iteration $t$, i.e., $t\in\mathcal I_{s_t}$.
In the algorithm and its analysis, $x_i^{(t)}$ denotes the local model of client $i$ at iteration $t$, while $\alpha_i^{(s)}$ denotes the $i$th component of the mixture vector $\alpha^{(s)}$ at round $s$.
Let $e_i\in \bbR^n$ denote the $i$th standard basis vector.
Algorithm~\ref{alg:fedavg_DRO} summarizes the procedure, whose two modules are detailed below.
\begin{algorithm}[htbp]
\small
\makeatletter
\long\def\@makecaption#1#2{\small\noindent{\bfseries #1} #2\par}
\makeatother
\algsetup{linenosize=\scriptsize}
\renewcommand{\algorithmicrequire}{\textbf{Input:}}
\renewcommand{\algorithmicensure}{\textbf{Output:}}
\setlength{\abovedisplayskip}{4pt plus 1pt minus 1pt}
\setlength{\belowdisplayskip}{4pt plus 1pt minus 1pt}
\setlength{\abovedisplayshortskip}{2pt plus 1pt}
\setlength{\belowdisplayshortskip}{2pt plus 1pt}
\hrule height 0.8pt\kern 2pt
\caption{BiDRO-FedAvg}
\label{alg:fedavg_DRO}
\kern 2pt\hrule height 0.4pt\kern 2pt
\begin{algorithmic}[1]
\REQUIRE
Number of clients $n$, communication interval $\tau$, total iterations $T$ (with $S:=T/\tau$ communication rounds),
learning rates $\eta_x,\eta_\alpha$, client sampling size $q$,
initial model $\bar{x}^{(0)}\in\mathcal{X}$ and initial weights $\alpha^{(0)}\in\Delta_{n-1}$.
\ENSURE
Final averaged solutions $\hat{x}_T=\frac{1}{qT}\sum_{t=0}^{T-1}\sum_{i\in\calC^{(s_t)}} x_i^{(t)}$,
$\hat{\alpha}_T=\frac{1}{S}\sum_{s=0}^{S-1}\alpha^{(s)}$.

\FOR{$s=0$ \textbf{to} $S-1$}
    \STATE Server samples a client batch $\calC^{(s)}=(c_1^{(s)},\ldots,c_q^{(s)})$ with \(\bbP(c_r^{(s)}=i\mid \alpha^{(s)})=\alpha_i^{(s)},
\quad i\in[n],\ r\in[q].\)
    \STATE Server independently samples $t^\prime \sim \mathrm{Unif}(\mathcal I_s)$
    \STATE Server broadcasts $\bar{x}^{(s)}$ and $t^\prime$ to all selected clients $i\in\calC^{(s)}$
    \FOR{client $i\in\calC^{(s)}$ \textbf{in parallel}}
        \STATE Set $x_i^{(s\tau)} \gets \bar{x}^{(s)}$
        \FOR{$t\in\mathcal I_s$}
            \STATE Obtain local sample $\widehat\zeta_i^{(t)}$ from $\empP{i}$
            \STATE Compute an $\epsilon$-approximate solution $z_i^{(t)}$ to $\max_{\xi\in\Xi}\; \{\ell(x_i^{(t)},\xi) - \rho_i\, c(\widehat{\zeta}_i^{(t)},\xi)\}$
            \STATE {Set} $x_i^{(t+1)} \gets \mathrm{Proj}_{\mathcal{X}}\!\Big(x_i^{(t)} - \eta_x \nabla_x \ell(x_i^{(t)},z_i^{(t)})\Big)$
        \ENDFOR
        \STATE Client sends $x_i^{((s+1)\tau)}$ and $x_i^{(t^\prime)}$ back to server
    \ENDFOR
    \STATE Server computes the average models:
    \[
        \bar{x}^{(s+1)} \gets \frac{1}{q}\sum_{i\in\calC^{(s)}} x_i^{((s+1)\tau)},
        \quad
        x^{(t^\prime)} \gets \frac{1}{q}\sum_{i\in\calC^{(s)}} x_i^{(t^\prime)}
    \]
    \STATE Server uniformly samples a subset \(\calU^{(s)}\!\subseteq\![n]\) of size \(q\)
    \STATE Server broadcasts $x^{(t^\prime)}$ to each client $i\in\mathcal{U}^{(s)}$
    \FORALL{clients $i\in\mathcal{U}^{(s)}$ \textbf{in parallel}}
        \STATE Obtain a fresh local sample $\widehat\zeta_i^{(t^\prime)}$ from $\empP{i}$, independently of the samples used in the primal updates
        \STATE Find an $\epsilon$-approximate maximizer $z_i^{(t^\prime)}$ of $\operatorname*{arg\,max}_{\xi\in\Xi}\; \ell(x^{(t^\prime)},\xi) - \rho_i\, c(\widehat{\zeta}_i^{(t^\prime)},\xi)$
        \STATE Client sends $\phi_i(x^{(t^\prime)},\widehat{\zeta}_i^{(t^\prime)},z_i^{(t^\prime)})$ back to server
    \ENDFOR
    \STATE Server forms $n$-dimensional vector $v\in\mathbb{R}^n$ by
    \[
        v \gets \frac{n}{q}\sum_{i\in\calU^{(s)}}
        \phi_i(x^{(t^\prime)},\widehat{\zeta}_i^{(t^\prime)},z_i^{(t^\prime)})e_i
    \]
    \STATE {Server updates the mixture weights}
    \[
        \alpha^{(s+1)} \gets \mathrm{Proj}_{\Delta_{n-1}}\!\big(\alpha^{(s)} + \eta_\alpha\, \tau v\big)
    \]
\ENDFOR
\end{algorithmic}
\kern 2pt\hrule height 0.4pt
\end{algorithm}

\textbf{Local model updates via adaptive sampling:} Let $\bar x^{(s)}$ and $\alpha^{(s)}$ be the global model and mixture weights at the server after communication round $s-1$.
At the beginning of the $s$-th round, the server draws a client batch $\calC^{(s)}=(c_1^{(s)},\ldots,c_q^{(s)})$, whose entries are sampled independently with replacement from \([n]\) according to the current mixture vector \(\alpha^{(s)}\). Consequently, the same client index may appear multiple times in \(\mathcal C^{(s)}\). Each time a client is selected, it runs a separate sequence of local updates initialized at $\bar x^{(s)}$, using fresh samples drawn independently across selections and local iterations. Repeated selections of the same client are counted separately in all sums and averages.
The server's current model $\bar{x}^{(s)}$ is broadcast to all selected clients $i\in\calC^{(s)}$.
With the global model, each client $i$ updates its local model by running $\tau$ iterations on its own data. 
Specifically, at each iteration $t$, each client $i$ draws a sample $\widehat{\zeta}_i^{(t)}\sim\empP{i}$ and solves the local problem:
\begin{align}\label{eq:inner_sup_xi}
    \operatorname*{arg\,max}_{\xi\in\Xi} \; \phi_i(x_i^{(t)},\widehat{\zeta}_i^{(t)},\xi)\,,
\end{align}
to obtain an $\epsilon$-approximate maximizer $z_i^{(t)}$ satisfying $\operatorname{dist}(z_i^{(t)},\calZ_i^{\star,(t)})\leq\epsilon$. Here, $\operatorname{dist}$ denotes the distance from a point to a set, and $\calZ_i^{\star,(t)}$ is the set of exact maximizers.
There could be a unique maximizer which can be compute efficiently when $\phi_i$ is strictly concave over $\xi$.
The approximate maximizer $z_i^{(t)}$ enables client $i$ to compute an approximate stochastic gradient with respect to $x$:
\begin{align}
    g_i^{(t)}&:=\nabla_{x}\ell(x_i^{(t)},z_i^{(t)})\,.\label{eq:gradient_x}
\end{align}
The resulting gradients are inexact, but the induced error is controlled by $\epsilon$ and will be accounted for in the convergence analysis. 
The local model is updated according to
\begin{equation}\label{eq:local_update}
    x_i^{(t+1)}=\mathrm{Proj}_{\calX}\left(x_i^{(t)}-\eta_xg_i^{(t)}\right)\,,
\end{equation}
where $\eta_x$ is the step size for updating the model parameter~$x$, and $\mathrm{Proj}_{\calX}(\cdot)$ denotes the projection on the set $\calX$ 
, i.e., $\mathrm{Proj}_{\calX}(y):=\arg\min_{x\in\calX}\|y-x\|$.
After $\tau$ local updates, client $i$ transmits $x_i^{((s+1)\tau)}$ to the server. The server aggregates the received local models to obtain the next global model
\begin{equation}
    \bar{x}^{(s+1)}=\frac{1}{q}\sum_{i\in\calC^{(s)}}x_i^{((s+1)\tau)}\,.
\end{equation}

\textbf{Mixture-weight update via snapshotting:}
The server keeps $\alpha^{(s)}$ fixed during the local model updates in communication round $s$ and updates it once at the end of that round.
Since the problem's objective is linear in $\alpha$, its gradient with respect to $\alpha$ depends only on the model $x$.
To form this round's mixture-weight update, we estimate the sum of dual gradients evaluated along the intermediate models in $\mathcal I_s$.
To estimate the sum without communicating at every local update iteration, we employ a random snapshotting scheme~\cite{deng2020distributionally}.
At the beginning of $s$th communication round, server samples a random iteration $t^\prime\in\mathcal I_s$ as the snapshot index, and sends it to selected clients in $\calC^{(s)}$.
The clients also return intermediate model $x_i^{(t^\prime)}$ along with the final model $x_i^{((s+1)\tau)}$ back to the server.
Similarly, the server computes an intermediate average model 
\begin{equation}
    x^{(t^\prime)}=\frac{1}{q}\sum_{i\in\calC^{(s)}}x_i^{(t^\prime)}\,,
\end{equation}
which is the basis of updating the mixture weights from $\alpha^{(s)}$ to $\alpha^{(s+1)}$.
Then the server selects a subset \(\calU^{(s)}\subseteq[n]\) of size \(q\), with a uniform distribution over all possible \(q\)-client subsets. Each selected client $i$ evaluates the following quantity:
\begin{equation}
\phi_i(x^{(t^\prime)},\widehat{\zeta}_i^{(t^\prime)},z_i^{(t^\prime)})
\label{eq:gradient_alpha}\,,
\end{equation}
where $z_i^{(t^\prime)}$ is an $\epsilon$-approximate maximizer of $\xi\mapsto\phi_i(x^{(t^\prime)},\widehat{\zeta}_i^{(t^\prime)},\xi)$ for each $i\in\calU^{(s)}$.
After receiving these values, the server constructs the vector $v$ as in Algorithm~\ref{alg:fedavg_DRO}, using
\[
v=\frac{n}{q}\sum_{i\in\calU^{(s)}}\phi_i(x^{(t^\prime)},\widehat{\zeta}_i^{(t^\prime)},z_i^{(t^\prime)})e_i\,.
\]
Finally, the mixture-weight update is given by
\begin{equation}\label{eq:update_alpha}
    \alpha^{(s+1)}=\mathrm{Proj}_{\Delta_{n-1}}\left(\alpha^{(s)}+\eta_\alpha \tau v\right)\,,
\end{equation}
where $\eta_\alpha$ is the step size for updating $\alpha$.
If the inner maximization is solved exactly, the snapshot construction makes $\tau v$ an unbiased estimator of the accumulated dual gradient within the communication round~\cite{deng2020distributionally}.
When only $\epsilon$-approximate inner maximizers are available, the estimator is no longer exact, but the induced deviation can be controlled through the approximation error.
\section{Convergence analysis}
We analyze Algorithm~\ref{alg:fedavg_DRO} with respect to the minimax problem~\eqref{eq:reformed_DRO}.
We impose the following assumptions. The regularity conditions in this section use the Euclidean norm.
\begin{assumption}\label{ass:main}
    \begin{enumerate}[(i)]
        \item The sets $\calX$ and $\Xi$ are nonempty, convex and compact, while $\calX$ has bounded diameter: $\sup_{x,x'\in\calX}\|x-x'\|\le D_{\calX}$.
        \item The loss function $\ell$ is continuous on $\calX\times\Xi$. For any $\xi\in\Xi$, $x\mapsto\ell(x,\xi)$ is convex and differentiable. For every $x\in\calX$, $\xi\mapsto\ell(x,\xi)$ is $L_{\xi}^{\ell}$-Lipschitz. Moreover, the mapping $\xi\mapsto \nabla_x\ell(x,\xi)$ is $L_{\xi}^{g}$-Lipschitz on $\Xi$. The primal gradient of the loss is uniformly bounded: there exists a constant $M_x^\ell>0$ such that
        \[
            \|\nabla_x\ell(x,\xi)\|\le M_x^\ell,
            \quad \forall x\in\calX,\ \xi\in\Xi .
        \]
        \item The transport cost $c$ is continuous on $\Xi\times\Xi$. For any $\zeta\in\Xi$, \(\xi\mapsto c(\zeta,\xi)\) is strongly convex and $L_{\xi}^{c}$-Lipschitz, with \(c(\zeta,\xi)\ge 0\) for all \(\zeta,\xi\in\Xi\).
        \item {The client-specific penalties $\rho_i>0$ are chosen such that, for every $i\in[n]$, $x\in\calX$, and $\zeta\in\Xi$, the mapping $\xi\mapsto\phi_i(x,\zeta,\xi)$ is strictly concave on $\Xi$.}
        \item For any $i\in[n]$, the local objective $F_i$ is $L_x^G$-smooth on $\calX$, i.e.,
        \[\|\nabla F_i(x)-\nabla F_i(x')\|\le L_x^G\|x-x'\|,\quad \forall x,x'\in\calX.\]
	\end{enumerate}
\end{assumption}
Assumption~\ref{ass:main}(i)--(iii) impose compactness, continuity, and transport-cost regularity conditions that are standard in Wasserstein DRO~\cite{mohajerinesfahani2018datadriven,sinha2017certifying}. Together with these conditions, Assumption~\ref{ass:main}(iv) ensures existence and uniqueness of the inner maximizer. The Lipschitz properties, upper bounds, the exact-oracle unbiasedness and moment bounds used in the convergence proof can be derived from the assumptions. The smoothness conditions imposed on the loss and the local robust objectives are also common in stochastic and federated minimax optimization~\cite{sinha2017certifying,deng2020distributionally}.

Considering multiple local updates on heterogeneous client objectives, a key step in the convergence analysis is to control the drift between local models {$x_i^{(t)}$} and their averaged counterpart, i.e., the virtual global model. To control the local-model drift uniformly over the admissible mixture weights, we adopt the following \emph{weighted gradient dissimilarity} $\Gamma\in\bbR_{\geq 0}$ from~\citet{deng2020distributionally}:
\begin{equation}\label{eq:Gamma_def}
\Gamma
:= \sup_{x\in\mathcal{X},\,\alpha\in\Delta_{n-1},\, i\in[n]}
\sum_{j=1}^n \alpha_j \bigl\|\nabla F_i(x)-\nabla F_j(x)\bigr\|^2 .
\end{equation}
The quantity \(\Gamma\) measures gradient heterogeneity among the local robust objectives and is zero if and only if all local gradients are identical on \(\calX\).

For the solution $(\hat{x}_T,\hat{\alpha}_T)\in\calX\times\Delta_{n-1}$ returned by Algorithm~\ref{alg:fedavg_DRO} {at the end of $T$ iterations,} we define the expected primal--dual gap as
\begin{equation}\label{eq:pd_gap_def}
    \operatorname{Gap}(\hat{x}_T,\hat{\alpha}_T)
    :=
    \max_{\alpha\in\Delta_{n-1}}
    \ev\big[F(\hat{x}_T,\alpha)\big]
    -
    \min_{x\in\calX}
    \ev\big[F(x,\hat{\alpha}_T)\big].
\end{equation}
We now establish the convergence result for the proposed algorithm, summarized in Theorem~\ref{thm:convergence}.
\begin{theorem}[Convergence of BiDRO-FedAvg]\label{thm:convergence}
    Let Assumption~\ref{ass:main} hold.
    Set $\tau=\max\{1,\lfloor T^{1/8}\rfloor\}$.
   If Algorithm~\ref{alg:fedavg_DRO} is run for \(T\) iterations with learning rates $\eta_x=\frac{1}{5L_{x}^{G}}T^{-5/8}$ and $\eta_\alpha=T^{-9/16}$, then the returned averaged solutions $(\hat{x}_T,\hat{\alpha}_T)$ satisfy
	    \begin{equation}\label{eq:theorem_gap}
	        {\operatorname{Gap}(\hat{x}_T,\hat{\alpha}_T)
	        =
	        \calO\left(T^{-3/8}+\epsilon\right).}
	    \end{equation}
	\end{theorem}
\proofdeferred{proof:convergence}
The iteration count $T=S\tau$ does not restrict the convergence rate. For any prescribed iteration budget \(\widetilde T\in\bbN_+\), we can set the parameters as follows:
\[
    \tau:=\max\{1,\lfloor\widetilde T^{1/8}\rfloor\},
    \qquad
    S:=\left\lfloor\frac{\widetilde T}{\tau}\right\rfloor,
    \qquad
    T:=S\tau.
\]
Then \(T\le\widetilde T\), \(T\) is divisible by \(\tau\), and fewer than \(\tau\) iterations of the prescribed budget remain unused. Moreover, \(\tau=\max\{1,\lfloor T^{1/8}\rfloor\}\) and \(T=\Theta(\widetilde T)\). Hence Theorem~\ref{thm:convergence} applies without changing the stated convergence rate.
\begin{remark}[Communication--Convergence Trade-Off]
    The parameter $\tau$ controls a communication--convergence trade-off. If one sets $\tau=T^b$ and chooses the corresponding stepsizes $\eta_x=\Theta(T^{-(1/2+b)})$ and $\eta_\alpha=\Theta(T^{-(1+b)/2})$, then the present analysis gives a rate of order $T^{-(1/2-b)}$ up to the sampling and approximation terms. A smaller $b$ means a smaller $\tau$ and a rate closer to the $T^{-1/2}$ limit associated with $\tau=1$; a larger $b$ reduces {$S$} but slows the provable convergence rate. The theorem fixes \(b=1/8\), i.e., \(\tau=\max\{1,\lfloor T^{1/8}\rfloor\}\), together with $\eta_x=\frac{1}{5L_{x}^{G}}T^{-5/8}$ and $\eta_\alpha=T^{-9/16}$, which yields the stated $T^{-3/8}$ dependence. \oprocend
\end{remark}
\begin{remark}[Role of Client Batch Size]
The parameters $\tau$ and $q$ play different roles in the convergence bound.
The parameter $\tau$ controls the local-drift error caused by multiple local updates, whereas \(q\) affects the sampling-dependent terms.
For a fixed \(\tau\), increasing \(q\) reduces the variance-related terms in the convergence bound, but it does not remove the local drift caused by client heterogeneity.
When \(\tau\) and \(q\) are tuned jointly, a larger \(q\) may lead to a better balance among sampling error, local drift, and communication cost.
Such an improvement, however, results from the joint parameter choice rather than from \(q\) alone. \oprocend
\end{remark}

\section{Numerical simulation}\label{sec:market_simulation}

We study a federated market-demand prediction problem constructed with both within-client distributional ambiguity and cross-client mixture uncertainty.
We evaluate three aspects of the proposed framework:
(i) the benefit of jointly modeling local and mixture uncertainty;
(ii) the performance of BiDRO-FedAvg when only a subset of clients performs multiple local updates in each round, under a communication-round budget;
and (iii) the sensitivity of BiDRO-FedAvg to varying local-distribution and client-mixture shifts, together with the algorithm's empirical convergence behavior.
\subsection{Experimental setup}
\begin{table*}[htbp]
\centering
\caption{Client construction in the market-demand experiment. The price and
advertising columns give the two shape parameters of the corresponding Beta
distribution.}
\label{tab:market_client_construction}
\scriptsize
\setlength{\tabcolsep}{4pt}
\begin{tabular}{clcccccc}
\toprule
$i$ & Market stage & Maturity interval {$[m_i^-,m_i^+]$}
& $(\kappa_{i,1}^\pi,\kappa_{i,2}^\pi)$
& $(\kappa_{i,1}^a,\kappa_{i,2}^a)$
& $\sigma_{y,i}$ & $N_i^{\rm tr}$ & $N_i^{\rm dep}$\\
\midrule
1 & Emerging      & $[0,.15)$   & $(2.0,5.0)$ & $(2.5,4.0)$ & $.025$ & 600 & 250\\
2 & Early growth  & $[.15,.30)$ & $(2.2,4.5)$ & $(3.0,3.5)$ & $.025$ & 150 & 500\\
3 & Growth        & $[.30,.50)$ & $(2.5,4.0)$ & $(3.5,3.0)$ & $.030$ & 100 & 750\\
4 & Mature        & $[.50,.70)$ & $(3.0,3.5)$ & $(3.0,3.0)$ & $.035$ & 75  & 1000\\
5 & Saturated     & $[.70,.85)$ & $(3.5,3.0)$ & $(2.5,3.5)$ & $.040$ & 50  & 1500\\
6 & Volatile late & $[.85,1]$   & $(4.0,2.5)$ & $(2.0,4.0)$ & $.045$ & 25  & 6000\\
\bottomrule
\end{tabular}
\end{table*}
We use $n=6$ clients representing, in increasing order of market maturity, emerging, early-growth, growth, mature, saturated, and volatile late markets. 
Let an observation be $\xi=(\theta,y)$, where
\[
    \theta=(m,\pi,a,1)\in\Theta,\quad \Theta:=[0,1]^3\times\{1\}\,,
\]
contains market maturity $m$, normalized price $\pi$, normalized advertising intensity $a$, and an intercept feature, while $y\in[0,1]$ is the realized demand.
The clients learn a shared model parameter $x\in\mathbb R^4$ and predict demand by
\[
    \widehat y=\theta^\top x,
    \qquad
    \ell(x,\xi)=(\theta^\top x-y)^2,
\]
with the feasible set $\mathcal X=\{x\in\mathbb R^4:\|x\|_2\le 1\}$.

We consider uncertainty in both the local data distributions and the client mixture weights. We first describe how the local training and deployment distributions are constructed.
Let ``$\mathrm{tr}$ ''and ``$\mathrm{dep}$'' denote the training and deployment environments, respectively.
The training samples define the empirical objective in problem~\eqref{eq:reformed_DRO}, and the model returned by BiDRO-FedAvg is evaluated on independently generated deployment samples.
For client $i$, environment $e\in\{\mathrm{tr},\mathrm{dep}\}$, and sample $k$, 
we independently generate the unshifted feature
    \[\bar\theta_{i,k}^e
    =(\bar m_{i,k}^e,\bar\pi_{i,k}^e,\bar a_{i,k}^e,1),\]
where $\bar{m}_{i,k}^e\sim\operatorname{Unif}[m_i^-,m_i^+],
    \bar{\pi}_{i,k}^e\sim
      \operatorname{Beta}(\kappa_{i,1}^\pi,\kappa_{i,2}^\pi)$, and $
    \bar{a}_{i,k}^e\sim
      \operatorname{Beta}(\kappa_{i,1}^a,\kappa_{i,2}^a)$.
The corresponding client-dependent parameters and sample sizes are reported in Table~\ref{tab:market_client_construction}. 
We next consider an environment-dependent feature offset \[s_i^e=(s_{i,m}^e,s_{i,\pi}^e,s_{i,a}^e,0),\] which shifts market maturity, price, and advertising while leaving the intercept unchanged.
The observed feature is then
\[
    \theta_{i,k}^e
    =(m_{i,k}^e,\pi_{i,k}^e,a_{i,k}^e,1)
    =\mathrm{Proj}_{[0,1]^3\times\{1\}}
      (\bar\theta_{i,k}^e+s_i^e).
\]
The underlying demand function is defined as
\[
\mu(\theta)=\mu(m,\pi,a):=h_0(m)+\beta_{\pi}(m)\pi+\beta_a(m)a,
\]
where functions $h_0$, $\beta_\pi$, and $\beta_a$ specify baseline demand, price sensitivity, and advertising effectiveness, respectively. They are piecewise linear in market maturity and are shared across all clients and both environments. Table~\ref{tab:demand_mechanism_parameters} gives their values at the maturity breakpoints $b_0,\ldots,b_6$.
For each $g\in\{h_0,\beta_\pi,\beta_a\}$, we use linear interpolation between adjacent breakpoints:
\[
 g(m)=\frac{b_{j+1}-m}{b_{j+1}-b_j}g(b_j)
      +\frac{m-b_j}{b_{j+1}-b_j}g(b_{j+1}),
\]
for $m\in[b_j,b_{j+1}]$ and $j=0,\ldots,5$. Since the feature projection ensures $m \in [0, 1]$, no extrapolation is needed.
\begin{table}[tbp]
\centering
\makeatletter
\renewcommand{\fnum@table}{\tablename~\thetable}
\makeatother
\caption{Breakpoint values of the structural demand functions.}
\label{tab:demand_mechanism_parameters}
\small
\begin{tabular}{rrrr}
\toprule
$b_j$ & $h_0(b_j)$ & $\beta_\pi(b_j)$ & $\beta_a(b_j)$\\
\midrule
$0$    & $.20$ & $-.10$ & $.30$\\
$.15$  & $.32$ & $-.18$ & $.36$\\
$.30$  & $.50$ & $-.28$ & $.42$\\
$.50$  & $.72$ & $-.36$ & $.34$\\
$.70$  & $.78$ & $-.42$ & $.22$\\
$.85$  & $.70$ & $-.35$ & $.16$\\
$1$    & $.62$ & $-.25$ & $.10$\\
\bottomrule
\end{tabular}
\end{table}

\begin{figure*}[t]
  \centering
  \includegraphics[width=0.8\textwidth]{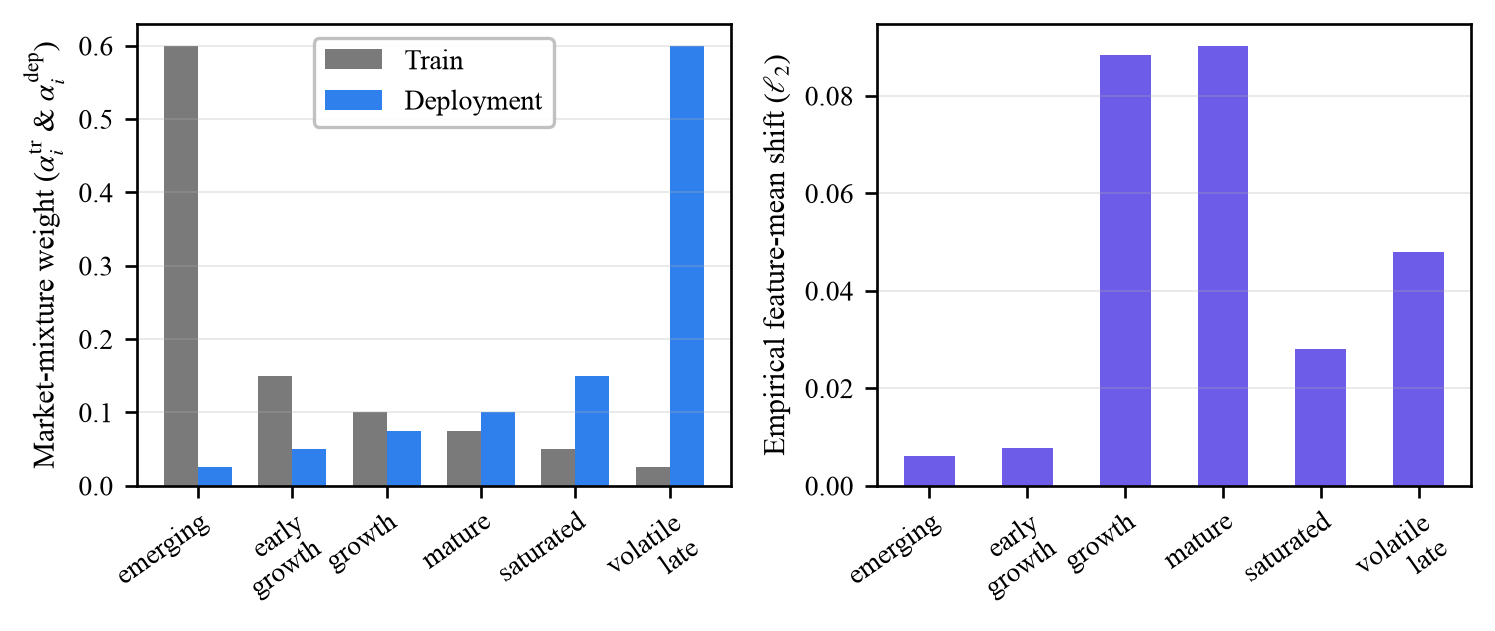}
  \caption{Training-to-deployment shifts in the market-demand experiment. The left panel shows the change from the early-market-heavy training mixture to the late-market-heavy deployment mixture. The right panel shows, for each client, the $\ell_2$ distance between the deployment and training sample means of the three features (market maturity, price, and advertising).}
  \label{fig:market_shift_diagnostics}
\end{figure*}

To reflect the complexity of demand relationships, we use a nonlinear data-generating mechanism while seeking a simple linear predictor that remains robust under distribution shifts. Our focus is therefore on predictive performance in changing environments rather than accurate recovery of the underlying demand function.
Conditional on this realized feature, demand is generated according to
\[
\begin{aligned}
    y_{i,k}^e
    =\mathrm{Proj}_{[0,1]}\!\Big(\mu(m_{i,k}^e,\pi_{i,k}^e,a_{i,k}^e)
      +\omega_{i,k}^e\Big)\,,
\end{aligned}
\]
where $\omega_{i,k}^e\sim\mathcal N(0,\sigma_{y,i}^2)$.
Client heterogeneity enters through the feature distributions and noise levels, as specified in Table~\ref{tab:market_client_construction}. We introduce the client-specific feature offsets below.
The final observation is given by
\[
    \xi_{i,k}^e=(\theta_{i,k}^e,y_{i,k}^e)\in\Xi,
    \qquad
    \Xi=\Theta\times[0,1].
\]
Applying this procedure independently gives the two finite datasets
$\mathcal D_i^{\rm tr}
    =\{\xi_{i,k}^{\rm tr}\}_{k=1}^{N_i^{\rm tr}},~
    \mathcal D_i^{\rm dep}
    =\{\xi_{i,k}^{\rm dep}\}_{k=1}^{N_i^{\rm dep}}$,
where $\mathcal D_i^{\rm tr}$ is used for learning, and the independently generated deployment dataset $\mathcal D_i^{\rm dep}$ is used to evaluate the prediction performance.
The training and deployment environments differ in both the client feature distributions and their mixture.
For client $i$, we vary the deployment offset according to
\begin{align}\label{eq:local_shift_path}
    {s_i^{\rm dep}(\lambda_{\mathrm{loc}})}
    =s_i^{\rm tr}+\lambda_{\mathrm{loc}}d_i,
    \qquad \lambda_{\mathrm{loc}}\in[0,1],
\end{align}
where $s_i^{\rm tr}$ is the training offset and $d_i$ is the deployment-drift direction, reported in Table~\ref{tab:market_feature_shifts}.
\begin{table}[tbp]
\centering
\makeatletter
\renewcommand{\fnum@table}{\tablename~\thetable}
\makeatother
\caption{Client-specific training offsets and deployment-drift
directions.}
\label{tab:market_feature_shifts}
\scriptsize
\setlength{\tabcolsep}{3pt}
\begin{tabular}{c@{\qquad}c@{\qquad}c}
\toprule
$i$ & $s_i^{\rm tr}$ & $d_i$\\
\midrule
1 & $(0,0,0,0)$       & $(0,0,0,0)$\\
2 & $(0,.01,-.01,0)$  & $(-.005,.010,-.010,0)$\\
3 & $(0,.02,-.01,0)$  & $(-.030,.035,-.035,0)$\\
4 & $(0,.03,-.02,0)$  & $(-.040,.045,-.045,0)$\\
5 & $(0,.04,-.03,0)$  & $(.020,-.025,.025,0)$\\
6 & $(0,.05,-.04,0)$  & $(.030,-.035,.035,0)$\\
\bottomrule
\end{tabular}
\end{table}
This offset changes the deployment feature distribution through the projection map above. At $\lambda_{\mathrm{loc}}=0$, the deployment feature distribution coincides with the training feature distribution. We set $\lambda_{\mathrm{loc}}=1$ as the default setting.
We next consider uncertainty in the client mixture weights, represented by a change from the training mixture $\alpha^{\rm tr}$ to the deployment mixture $\alpha^{\rm dep}$.
The training sample sizes in Table~\ref{tab:market_client_construction} induce the early-market-heavy empirical mixture
$\alpha^{\mathrm{tr}}
    =(0.60,0.15,0.10,0.075,0.05,0.025)$.
The default deployment mixture is
\[
    \alpha^{\mathrm{dep}}
    =(0.025,0.05,0.075,0.10,0.15,0.60),
\]
which assigns most mass to the volatile late-market client. 
We parameterize the mixture shift by
\[
    \alpha(\lambda_{\mathrm{mix}})
    =(1-\lambda_{\mathrm{mix}})\alpha^{\mathrm{tr}}
      +\lambda_{\mathrm{mix}}\alpha^{\mathrm{dep}},
\]
where $\lambda_{\mathrm{mix}}\in[0,1]$ controls the mixture-shift intensity.
Hence, $\lambda_{\mathrm{mix}}=0$ and $\lambda_{\mathrm{mix}}=1$ correspond to the training and default deployment mixtures, respectively.
Unless otherwise stated, we set $(\lambda_{\mathrm{loc}}, \lambda_{\mathrm{mix}}) = (1, 1)$. The experiment combines a global client-mixture shift, $\alpha^{\rm tr}\mapsto\alpha^{\rm dep}$, with client-wise feature-distribution shifts induced by the deployment offsets, $\bbP_i^{\rm tr}\mapsto\bbP_i^{\rm dep}$. Figure~\ref{fig:market_shift_diagnostics} presents these changes from left to right: the left panel shows the training and deployment client-mixture weights, and the right panel shows the empirical $\ell_2$ shift between the training and deployment sample means of the three features for each client.
For sensitivity analysis, we vary $\lambda_{\mathrm{mix}}\in\{0,0.125,0.25,0.375,0.5,0.625,0.75,0.875,1\}$,
while fixing $\lambda_{\mathrm{loc}}=1$, and vary $\lambda_{\mathrm{loc}}\in\{0,0.25,0.5,0.75,1\}$ while fixing the client mixture at $\alpha^{\mathrm{dep}}$.
Each sweep varies one source of distribution shift while holding the other at the default value.

We compare BiDRO-FedAvg with FedAvg~\cite{mcmahan2017communication}, AFL~\cite{mohri2019agnostica}, DRFA~\cite{deng2020distributionally}, and DRO-FL \cite{wang2025distributionally}. FedAvg aggregates clients according to their training sample proportions. AFL and DRFA optimize worst-case client-mixture weights but evaluate local losses on empirical samples.
For DRO-FL and BiDRO-FedAvg, the empirical objective function is
\[
\begin{aligned}
    F_i(x)
    &=\frac{1}{N_i^{\rm tr}}
      \sum_{k=1}^{N_i^{\rm tr}}f_i(x,\xi_{i,k}^{\rm tr})\,,
\end{aligned}
\]
where $f_i(x,\zeta)=\sup_{\xi\in\Xi}\{\ell(x,\xi)-\rho_i\|\xi-\zeta\|_2^2\}$.
DRO-FL uses the common penalty $\rho_i=8$ for all clients, whereas
BiDRO-FedAvg uses 
\[(\rho_1,\ldots,\rho_6)=(9.6,8.8,7.2,6.8,6.0,5.6),\] where smaller penalties correspond to larger effective ambiguity regions.
Since the resulting inner problems are strictly concave in $\xi$ under the implementation, we solve inner problems KKT equations, using at most $30$ steps with tolerance $10^{-7}$.
We run $T=3^8=6561$ iterations and repeat each experiment over $10$ paired data realizations. Unless otherwise stated, we use $q=4$ and $\tau=3$.

We use two complementary protocols since the compared methods have different values of $\tau$.
Under \emph{full-participation protocol}, we set $q=n$, $\tau=1$, and $T=6561$. Every client participates, and the server aggregates after every local update. This protocol removes the effects of client subsampling and multiple local steps, allowing the ambiguity models to be compared under $q=n$ and $\tau=1$.
Under the \emph{matched-round protocol}, all methods use the same $S$. In this protocol, FedAvg, DRFA, and BiDRO-FedAvg use $q=4$, $\tau=3$, and $T=3^8=6561$, giving $S=T/\tau=2187$. AFL and DRO-FL use $\tau=1$ and hence $T=S=2187$.  For DRFA and BiDRO-FedAvg, at each of their communication rounds, the $q$ selected clients are sampled independently with replacement according to the current adversarial mixture.
For client $i$, define its empirical deployment MSE as
\[
R_i(x)
=\frac{1}{N_i^{\rm dep}}
  \sum_{k=1}^{N_i^{\rm dep}}
  \left((\theta_{i,k}^{\rm dep})^\top x-y_{i,k}^{\rm dep}\right)^2.
\]
The deployment-average MSE is calculated by
\[
R_{\mathrm{avg}}(x)=\sum_{i=1}^n\alpha_i^{\mathrm{dep}}R_i(x).
\]
Finally, to examine the empirical  convergence behavior of the proposed algorithm, at each horizon $T$, we apply the stepsizes $\eta_x=0.6T^{-5/8}$ and $\eta_\alpha=1.4T^{-9/16}$,
and evaluate the ergodic output $\widehat x=\bar x$ returned by Algorithm~\ref{alg:fedavg_DRO}.

\subsection{Results}
\begin{figure*}[t]
  \centering
  \begin{subfigure}{0.42\textwidth}
    \centering
    \includegraphics[width=\linewidth]{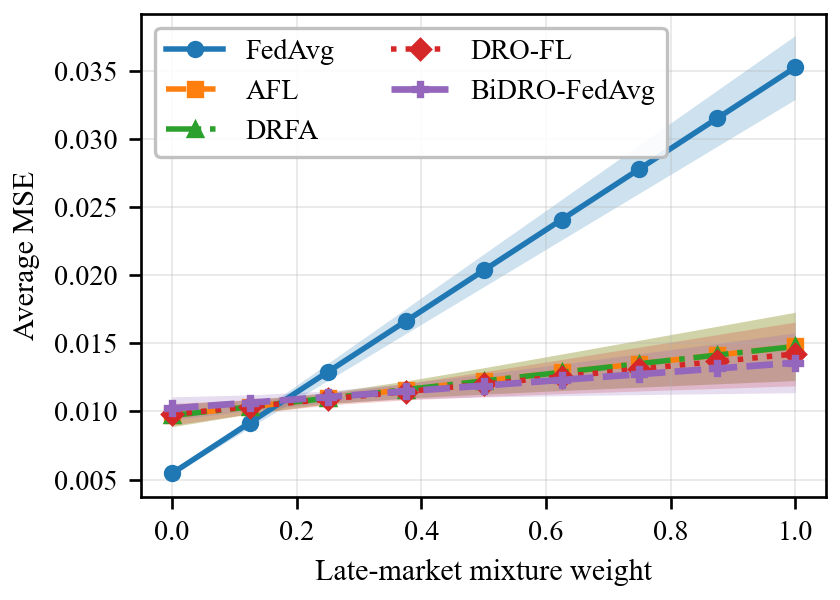}
    \caption{Full participation: mixture shift.}\label{fig:3a}
  \end{subfigure}
  \hfill
  \begin{subfigure}{0.42\textwidth}
    \centering
    \includegraphics[width=\linewidth]{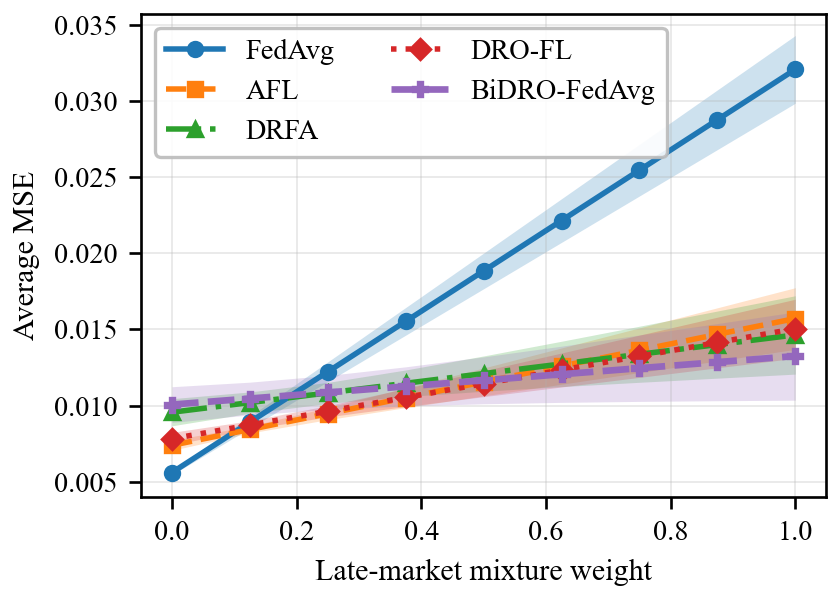}
    \caption{Matched rounds: mixture shift.}\label{fig:3b}
  \end{subfigure}
  \\[1mm]
  \begin{subfigure}{0.42\textwidth}
    \centering
    \includegraphics[width=\linewidth]{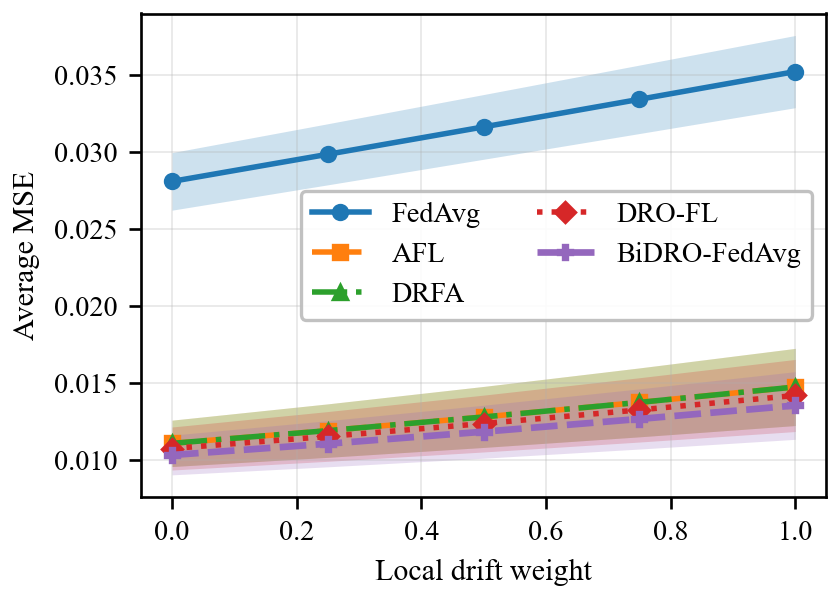}
    \caption{Full participation: local drift.}\label{fig:3c}
  \end{subfigure}
  \hfill
  \begin{subfigure}{0.42\textwidth}
    \centering
    \includegraphics[width=\linewidth]{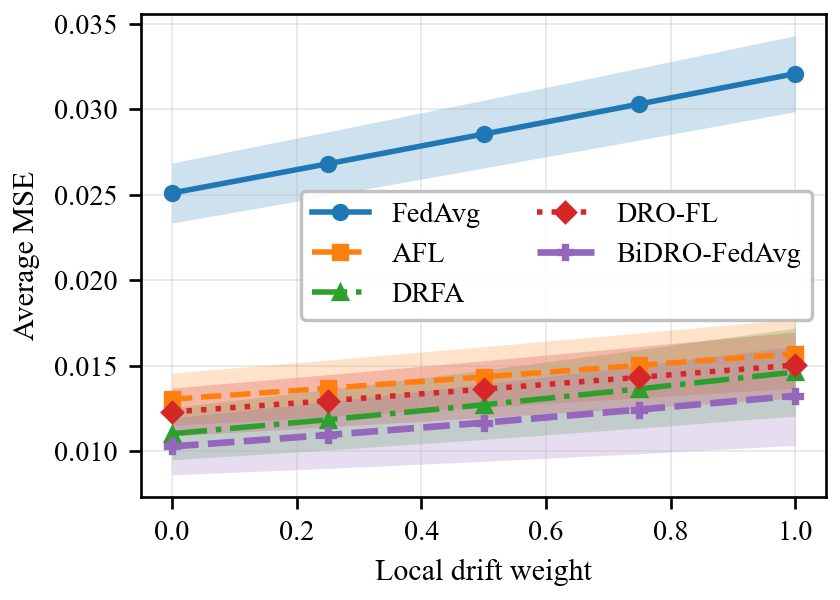}
    \caption{Matched rounds: local drift.}\label{fig:3d}
  \end{subfigure}
  \caption{Deployment-average MSE under varying deployment shifts and communication protocols. The top row varies the client-mixture shift intensity $\lambda_{\mathrm{mix}}$, whereas the bottom row varies the local feature-distribution shift intensity $\lambda_{\mathrm{loc}}$. The left column reports results under full participation, and the right column reports results under the matched-round protocol. Curves show averages over $10$ repetitions, with shaded regions indicating one standard deviation.}
  \label{fig:market_sweeps}
\end{figure*}
We first report performance in the default environment.
Table~\ref{tab:market_results} summarizes the deployment MSE under the full-participation and matched-round protocols.
We begin with the full-participation protocol. BiDRO-FedAvg achieves the lowest deployment MSE of $0.0136$, compared with $0.0148$ for DRFA and $0.0142$ for homogeneous DRO-FL. Under this common communication pattern, the improvement over DRFA supports the benefit of incorporating client-wise local robustness in addition to global mixture robustness. BiDRO-FedAvg also attains a lower mean than DRO-FL with a common penalty, which is consistent with the potential benefit of allowing client-dependent robustness levels. These comparisons therefore support the use of our proposed ambiguity models.
We next consider the matched-round protocol. As reported in Table~\ref{tab:market_results}, BiDRO-FedAvg achieves a deployment MSE of $0.0133$, compared with $0.0146$ for DRFA, $0.0150$ for DRO-FL, and $0.0157$ for AFL. Thus, the advantage of BiDRO-FedAvg is retained when only selected clients participate and multiple local steps are performed between aggregations. This result demonstrates its prediction efficiency under the same budget of communication rounds.
\begin{table}[htbp]
\centering
\makeatletter
\renewcommand{\fnum@table}{\tablename~\thetable}
\makeatother
\caption{Deployment MSE under the full-participation and matched-round protocols in the default coupled-shift environment. Entries are mean $\pm$ standard deviation over $10$ paired repetitions. The matched-round protocol uses $2187$ communication rounds.}
\label{tab:market_results}
\scriptsize
\setlength{\tabcolsep}{2.2pt}
\begin{tabular}{lcc}
\toprule
Method & Full participation & Matched rounds\\
\midrule
FedAvg & $0.0352\pm0.0023$ & $0.0321\pm0.0022$\\
AFL & $0.0148\pm0.0025$ & $0.0157\pm0.0020$\\
DRFA & $0.0148\pm0.0025$ & $0.0146\pm0.0026$\\
DRO-FL & $0.0142\pm0.0023$ & $0.0150\pm0.0019$\\
BiDRO-FedAvg & $\mathbf{0.0136\pm0.0022}$ & $\mathbf{0.0133\pm0.0029}$\\
\bottomrule
\end{tabular}
\end{table}

Fig.~\ref{fig:market_sweeps} examines performance as one source of deployment shift varies while the other remains fixed at its default value.
In Fig.~\ref{fig:3a} and Fig.~\ref{fig:3b}, FedAvg has the lowest deployment-average MSE when $\lambda_{\mathrm{mix}}=0$. 
As the mixture shifts toward late-market clients, BiDRO-FedAvg achieves the lowest average MSE from $\lambda_{\mathrm{mix}}=0.5$ under full participation and from $\lambda_{\mathrm{mix}}=0.625$ under the matched-round protocol. In the local-drift sweep, all methods deteriorate as the drift increases, while BiDRO-FedAvg has the lowest average MSE at all five levels under both protocols. These results show that the advantage is not restricted to the default deployment environment.

\begin{figure}[!htbp]
  \centering
  \includegraphics[width=0.8\linewidth]{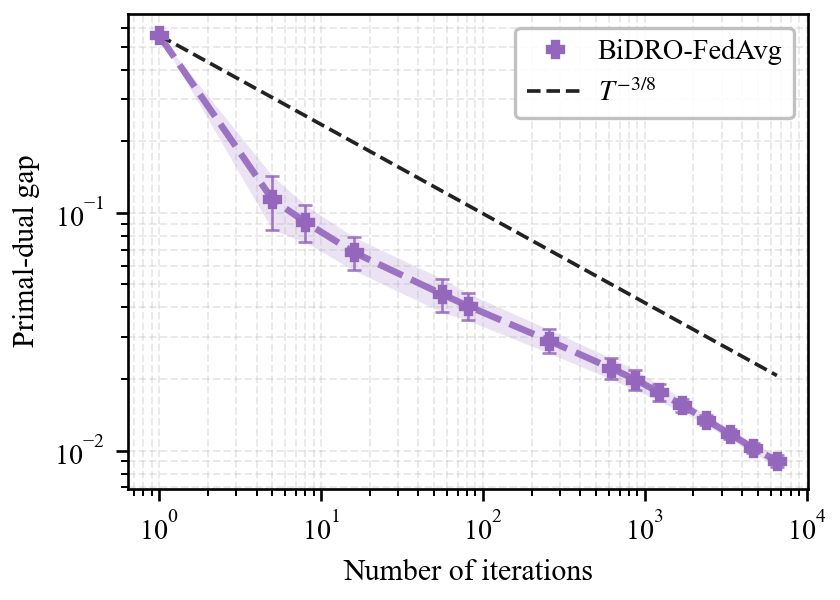}
  \caption{The primal--dual gap of BiDRO-FedAvg across iteration budgets with  a reference line proportional to $T^{-3/8}$. Markers and error bars show the mean and one standard deviation over 10 repetitions.}
  \label{fig:theorem_convergence_simulation}
\end{figure}
Finally, we move to the convergence behavior of BiDRO-FedAvg. Fig.~\ref{fig:theorem_convergence_simulation} compares the realized primal--dual gap with the $T^{-3/8}$ rate predicted by Theorem~\ref{thm:convergence}. The gap decreases from $0.5583$ at $T=1$ to $0.0090$ at $T=6561$, while its variation across repetitions also contracts. A log--log fit over the eight largest iteration budgets yields a slope of $-0.388$, close to the theoretical exponent $-3/8=-0.375$. The observed decay is therefore consistent with the convergence order established in
Theorem~\ref{thm:convergence}.

Overall, the simulations support the effectiveness of the proposed modeling and algorithm. Under full participation, jointly accounting for cross-client mixture uncertainty and within-client local uncertainty improves prediction performance in shifted deployment environments. 
BiDRO-FedAvg retains this advantage under partial participation and exhibits convergence consistent with the theoretical rate.

\section{Conclusion}\label{sec:conclusion}
This paper addresses joint uncertainty in local data distributions and deployment mixture weights through admissible mixtures of client-wise Wasserstein ambiguity sets.
The formulation allows different levels of local robustness while preserving client-wise separability, and admits high-probability coverage and out-of-sample guarantees.
We develop BiDRO-FedAvg for a penalized formulation with an induced-radius robustness interpretation.
The algorithm supports partial client participation and multiple local updates, with a convergence guarantee.
In the default market-demand setting with both local feature-distribution and client-mixture shifts, BiDRO-FedAvg achieved the lowest mean deployment MSE among the compared methods under both full-participation and matched-round protocols.
The observed primal--dual gap decayed consistently with the theoretically predicted rate.

\section*{Declaration of generative AI and AI-assisted technologies in the writing process}
The authors used ChatGPT (OpenAI) to improve language and assist in checking mathematical derivations. They reviewed and edited all AI-assisted content, independently verified the mathematical arguments, and take full responsibility for the manuscript.

\ifdefined\DROFullVersion
  \begin{center}
  {\LARGE Appendix\par}
  \end{center}
  \appendix
  \numberwithin{theorem}{section}
  \section{Proofs of out-of-sample performance guarantees}
\subsection{Proof of Proposition~\mref{prop:coverage}}\label{proof:coverage}
 
We first show that $\Pr\!\left(\bbP\in\calW_{\mathrm G}\right)\ge \prod_{i=1}^n(1-\beta_i)$. For each client $i$, define the event
\[
E_i:=\bigl\{\bbP_i\in \calW_i(\varepsilon_i)\bigr\}
=\bigl\{W_p(\bbP_i,\empP{i})\le \varepsilon_i\bigr\}.
\]
By Assumption~\mref{ass:local_coverage}, $\bbP_i^{\otimes N_i}\!\left(E_i\right)\ge 1-\beta_i$ for all $i$.
Since the local datasets are mutually independent across clients, the events $\{E_i\}_{i=1}^n$
are mutually independent, and thus
\[
\Pr\!\left(\bigcap_{i=1}^n E_i\right)
=\prod_{i=1}^n {\bbP_i^{\otimes N_i}}\!\left(E_i\right)
\ge \prod_{i=1}^n(1-\beta_i).
\]
On the event $\bigcap_{i=1}^n E_i$, we have $\bbP_i\in\calW_i(\varepsilon_i)$ for all $i$.
Together with the mixture representation of $\bbP$ in~\meqref{eq:mixture4ground_truth},
we have $\bbP\in\calW_{\mathrm G}$ due to the definition of $\calW_{\mathrm G}$ in~\meqref{eq:global_ambiguity_set}.
Therefore,
\[
\Pr\!\left(\bbP\in\calW_{\mathrm G}\right)\ge \Pr\!\left(\bigcap_{i=1}^n E_i\right)\ge \prod_{i=1}^n(1-\beta_i).
\]
We now prove the second bound. For each client \(i\), define the intermediate event
\[
E_i^\sharp
:=
\begin{cases}
\Bigl\{\,W_p(\bbP_i,\empP{i})\le \varepsilon_i-\delta\,\Bigr\},
& \varepsilon_i\ge\delta,\\
\emptyset, & \varepsilon_i<\delta.
\end{cases}
\]
By Assumption~\mref{ass:local_coverage} and the definition of $\bar\beta_i$, we have
\begin{align*}
    \Pr\!\left(E_i^\sharp\right)\ge 1-\bar\beta_i,
\end{align*}
where $\bar\beta_i$ bounds the failure probability of the strengthened local containment event $E_i^\sharp$.
Next, define
\[
\tilde E_i
:=
\Bigl\{\,W_p(\bbP_j,\empP{i})\le \varepsilon_i,\ \forall j\in[n]\,\Bigr\},
\]
that is, the local ambiguity ball centered at \(\empP{i}\) contains all true local distributions.
On the event \(E_i^\sharp\), Assumption~\mref{ass:bounded_heterogeneity} and the triangle inequality imply that, for every \(j\in[n]\), 
\begin{align}
    W_p(\bbP_j,\empP{i})
\ 
\le&~
W_p(\bbP_i,\empP{i})+W_p(\bbP_i,\bbP_j)
\ \notag\\
\le&~
W_p(\bbP_i,\empP{i})+\delta.
\end{align}
Hence, on the event \(E_i^\sharp\), we have $W_p(\bbP_j,\empP{i})\le \varepsilon_i$, $\forall j\in[n]$, i.e., 
$E_i^\sharp\subseteq\tilde{E}_i$, which
indicates that $\Pr\!\left(\tilde{E}_i\right)\geq\Pr\!\left(E_i^\sharp\right)\geq 1-\bar{\beta}_i$.
For any $i\in [n]$, since \(\bbP\) is a convex combination of \(\{\bbP_i\}_{i=1}^n\), the event $\tilde E_i$ implies \(\{\bbP\in\calW_{\mathrm G}\}\).
Since it suffices that \emph{there exists} such a representative client, we have
\[
\Pr\!\left(\bbP\in\calW_{\mathrm G}\right)
\ge
\Pr\!\left(\bigcup_{i=1}^n \tilde E_i\right)
\ge
\Pr\!\left(\bigcup_{i=1}^n E_i^\sharp\right).
\]
Finally, because the local datasets are independent across clients, the events $\{E_i^\sharp\}_{i=1}^n$ are mutually independent. This yields
\[
\Pr\!\left(\bigcup_{i=1}^n E_i^\sharp\right)
=
1-\prod_{i=1}^n \Pr\!\left((E_i^\sharp)^c\right)
\ge
1-\prod_{i=1}^n \bar\beta_i,
\]
completing the proof.

\subsection{Proof of Lemma~\mref{lm:separability}}\label{proof:separability}
 
    For any fixed $x \in \calX$, the union structure of $\calW_{\mathrm G}$ yields:
    \begin{align}\label{eq:separable_sup}
        \max_{\bbQ\in\calW_{\mathrm G}}\ev_{\bbQ}[\ell(x,\xi)]
        &= \max_{\bbQ\in\cup_{\alpha\in\Delta_{n-1}}\calW_{\mathrm M}(\alpha)}
        \ev_{\bbQ}[\ell(x,\xi)]\notag \\
        &= \max_{\alpha\in\Delta_{n-1}}
        \max_{\bbQ\in\calW_{\mathrm M}(\alpha)}
        \ev_{\bbQ}[\ell(x,\xi)].
    \end{align}
    For any fixed $\alpha\in\Delta_{n-1}$, following the decomposition argument in \citet[Proposition~1]{ibrahim2025fdrsvm}, we have
\begin{align}
\max_{\bbQ\in\calW_{\mathrm M}(\alpha)}
\ev_{\bbQ}[\ell(x,\xi)]
&=\max_{\{\bbQ_i\in\calW_i(\varepsilon_i)\}_{i=1}^n}
\ev_{\sum_{i=1}^n\alpha_i\bbQ_i}[\ell(x,\xi)] \notag\\
&=\max_{\{\bbQ_i\in\calW_i(\varepsilon_i)\}_{i=1}^n}
\sum_{i=1}^n\alpha_i\ev_{\bbQ_i}[\ell(x,\xi)]\notag\\
&=\sum_{i=1}^n\alpha_i
\max_{\bbQ_i\in\calW_i(\varepsilon_i)}
\ev_{\bbQ_i}[\ell(x,\xi)]\,,
\label{eq:fixed_alpha_separability}
\end{align}
where the first equality follows from the definition of $\calW_{\mathrm M}(\alpha)$ in~\meqref{eq:global_ambiguity}, the second equality is a result of applying the law of total expectation for the mixture distribution \(\bbQ=\sum_{i=1}^n \alpha_i \bbQ_i\), and the last equality follows from the separability of the maximization, since each $\bbQ_i$ only affects its corresponding summand.
Substituting~\eqref{eq:fixed_alpha_separability} into~\eqref{eq:separable_sup} and minimizing over $x\in\calX$ completes the proof.

\subsection{Proof of Theorem~\mref{thm:OOS_separated}}\label{proof:oos}
To prove the theorem, we use the following auxiliary inequality.
 \begin{lemma}\label{lm:inequality}
    Let $\alpha^{\mathrm{I}},\alpha^{\mathrm{II}}\in\Delta_{n-1}$ and $b\in\bbR^n$. Then, the following inequality holds:
    \begin{align}
        |(\alpha^{\mathrm{I}}-\alpha^{\mathrm{II}})^\top b|\leq\frac{1}{2}\|\alpha^{\mathrm{I}}-\alpha^{\mathrm{II}}\|_1\Bigl(\max_{i}b_i-\min_i b_i\Bigr)\,.
    \end{align}
\end{lemma}
\begin{proof}
    Since $\alpha^{\mathrm{I}},\alpha^{\mathrm{II}}\in\Delta_{n-1}$, we have $\sum_{i=1}^n(\alpha^{\mathrm{I}}_i-\alpha^{\mathrm{II}}_i)=0$.  Thus, for any $c\in\bbR$, we have
\begin{align}
            (\alpha^{\mathrm{I}}-\alpha^{\mathrm{II}})^\top b&=\sum_{i=1}^n(\alpha^{\mathrm{I}}_i-\alpha^{\mathrm{II}}_i)b_i-c\sum_{i=1}^n(\alpha^{\mathrm{I}}_i-\alpha^{\mathrm{II}}_i)\notag\\&=\sum_{i=1}^n(\alpha^{\mathrm{I}}_i-\alpha^{\mathrm{II}}_i)(b_i-c).\label{eq:lm_inequality}
\end{align}
    By choosing $c=(\max_ib_i+\min_ib_i)/{2}$, the inequality
    \begin{align}\label{eq:lm_inequality_2}
        |b_i-c|\leq \frac{1}{2}(\max_ib_i-\min_ib_i)
    \end{align}
    holds for $\forall i\in[n]$.
    As a consequence, 
taking the modulus
 on both sides of~\eqref{eq:lm_inequality} leads to
\begin{align}
             |(\alpha^{\mathrm{I}}-\alpha^{\mathrm{II}})^\top b|&=\Bigr|\sum_{i=1}^n(\alpha^{\mathrm{I}}_i-\alpha^{\mathrm{II}}_i)(b_i-c)\Bigr|\notag\\
          &   \leq \sum_{i=1}^n|\alpha^{\mathrm{I}}_i-\alpha^{\mathrm{II}}_i|\cdot|b_i-c|\notag\\
          &\leq \frac{1}{2}(\max_ib_i-\min_ib_i)\sum_{i=1}^n|\alpha^{\mathrm{I}}_i-\alpha^{\mathrm{II}}_i|
\end{align}
    where the last inequality follows from~\eqref{eq:lm_inequality_2}.
\end{proof}

We now prove the theorem. The first statement follows directly from Proposition~\mref{prop:coverage}.
On the event $\{\bbP\in\calW_{\mathrm G}\}$ which occurs with probability at least $1-\beta$, we have
\[
\Josp=\ev_{\bbP}[\ell(\xdro,\xi)]
\le \sup_{\bbQ\in\calW_{\mathrm G}}\ev_{\bbQ}[\ell(\xdro,\xi)]
= \Jdro,
\]
where we recall that $\xdro$ is an optimal solution
of \meqref{eq:problem_DRO}.
This proves~\meqref{eq:oos_upper_bound}.

For the second statement, to bound $\Jdro-\Josp$, we select a worst-case weight vector $\widehat{\alpha}$ and an associated worst-case distribution $\bbQ_{\calW}$:
\begin{subequations}
    \begin{align}
        \widehat{\alpha}&\in\operatorname*{arg\,max}\limits_{\alpha\in\Delta_{n-1}}\max_{\bbQ\in\calW_{\mathrm M}(\alpha)}
\ev_{\bbQ}[\ell(\xdro,\xi)]\,,\\
        \bbQ_\calW&\in\operatorname*{arg\,max}\limits_{\bbQ\in\calW_{\mathrm M}(\widehat{\alpha})}\mathbb{E}_{\bbQ}[\ell(\xdro,\xi)]\,.
    \end{align}
\end{subequations}
The existence of these maximizers follows from compactness and continuity, as detailed below.
Since $\Xi$ is compact, Prokhorov's theorem~\cite{dudley2018real} implies that the space of probability measures on $\Xi$ is weakly compact.
Each local Wasserstein ball $\calW_i(\varepsilon_i)$ is nonempty, as it contains $\empP{i}$, and weakly closed, and is therefore weakly compact.
For every fixed $\alpha\in\Delta_{n-1}$, the mixing map
$(\bbQ_1,\ldots,\bbQ_n)\mapsto\sum_{i=1}^n\alpha_i\bbQ_i$
is continuous under weak convergence.
Thus, $\calW_{\mathrm M}(\alpha)$ is weakly compact as the image of the compact set $\prod_{i=1}^n\calW_i(\varepsilon_i)$.
Since $\ell(\xdro,\cdot)$ is bounded and continuous, the expected-loss functional is weakly continuous and attains its maximum over both the local balls and their fixed-weight mixture sets.
Moreover, by~\eqref{eq:fixed_alpha_separability}, the inner optimal value is linear in $\alpha$ and hence attains its maximum on the compact simplex $\Delta_{n-1}$.
By separability, we may choose these maximizers so that
$\bbQ_{\calW}=\sum_{i=1}^n\widehat{\alpha}_i\bbQ_{\calW}^{(i)}$,
where, for each client $i$, we select a local worst-case distribution
\[
\bbQ_{\calW}^{(i)}\in\operatorname*{arg\,max}_{\bbQ_i\in\calW_i(\varepsilon_i)}
\ev_{\bbQ_i}[\ell(\xdro,\xi)].
\]
Thus, the performance gap can be decomposed as
\begin{align}
        |\Jdro&~-\Josp|
        =|\ev_{\bbQ_\calW}[\ell(\xdro,\xi)]-\ev_\bbP[\ell(\xdro,\xi)]|\notag\\
        =&~|\sum_{i=1}^n\widehat{\alpha}_i \ev_{\bbQ_{\calW}^{(i)}}[\ell(\xdro,\xi)]-\sum_{i=1}^n\alpha_i^\star\ev_{\bbP_i}[\ell(\xdro,\xi)]|\notag\\
        =&~\Bigl|\sum_{i=1}^n\widehat{\alpha}_i \ev_{\bbQ_{\calW}^{(i)}}[\ell(\xdro,\xi)]-\sum_{i=1}^n\widehat{\alpha}_i\ev_{\bbP_i}[\ell(\xdro,\xi)]\notag\\
        &+\sum_{i=1}^n\widehat{\alpha}_i\ev_{\bbP_i}[\ell(\xdro,\xi)]-\sum_{i=1}^n\alpha_i^\star\ev_{\bbP_i}[\ell(\xdro,\xi)]\Bigr|\notag\\
        \leq&~ \underbrace{\Bigl|\sum_{i=1}^n\widehat{\alpha}_i \left(\ev_{\bbQ_{\calW}^{(i)}}[\ell(\xdro,\xi)]-\ev_{\bbP_i}[\ell(\xdro,\xi)]\right)\Bigr|}_{\text{sampling}}\notag\\
        &+\underbrace{\Bigl|\sum_{i=1}^n(\widehat{\alpha}_i-\alpha_i^\star)\ev_{\bbP_i}[\ell(\xdro,\xi)]\Bigr|}_{\text{ heterogeneity \& weights}}\,.\label{eq:OOS_separation}
\end{align}
We next bound these two resulting terms.
If $L_\xi^\ell=0$, then $\ell(x,\cdot)$ is constant for each $x$, so $\Jdro=\Josp$ and the second statement is immediate.
Since $\ell(x,\cdot)$ is $L_\xi^\ell$-Lipschitz,
the normalized loss $\ell(x,\cdot)/L_\xi^\ell$ is $1$-Lipschitz
for every fixed $x$, provided $L_\xi^\ell>0$.
Let $\|h\|_{\mathrm{Lip}}$ denote the Lipschitz constant
of an arbitrary test function $h:\Xi\to\bbR$ with respect to the ground metric $d$.
Consider the event $\bigcap_{i=1}^n E_i$, on which
we obtain the following bounds for the sampling term:
\begin{align}\label{eq:bound1}
&\Bigl|
\sum_{i=1}^n\widehat{\alpha}_i
\left(
\ev_{\bbQ_{\calW}^{(i)}}[\ell(\xdro,\xi)]
-\ev_{\bbP_i}[\ell(\xdro,\xi)]
\right)
\Bigr| \notag\\
\leq{}&
\sum_{i=1}^n\widehat{\alpha}_i
\Bigl|
\ev_{\bbQ_{\calW}^{(i)}}[\ell(\xdro,\xi)]
-\ev_{\bbP_i}[\ell(\xdro,\xi)]
\Bigr| \notag\\
={}&
L_\xi^\ell\sum_{i=1}^n\widehat{\alpha}_i
\Bigl|
\ev_{\bbQ_{\calW}^{(i)}}\Bigl[\frac{1}{L_\xi^\ell}\ell(\xdro,\xi)\Bigr]
-\ev_{\bbP_i}\Bigl[\frac{1}{L_\xi^\ell}\ell(\xdro,\xi)\Bigr]
\Bigr| \notag\\
\overset{\mathrm{(a)}}{\leq}{}&
L_\xi^\ell\sum_{i=1}^n\widehat{\alpha}_i
\sup_{\|h\|_{\mathrm{Lip}}\leq1}
\Bigl|
\ev_{\bbQ_{\calW}^{(i)}}[h(\xi)]
-\ev_{\bbP_i}[h(\xi)]
\Bigr| \notag\\
\overset{\mathrm{(b)}}{=}{}&
L_\xi^\ell\sum_{i=1}^n\widehat{\alpha}_i
W_1(\bbQ_{\calW}^{(i)},\bbP_i) \notag\\
\overset{\mathrm{(c)}}{\leq}{}&
L_\xi^\ell\sum_{i=1}^n\widehat{\alpha}_i
W_p(\bbQ_{\calW}^{(i)},\bbP_i) \notag\\
\leq{}&
L_\xi^\ell\sum_{i=1}^n\widehat{\alpha}_i
\Bigl(
W_p(\bbQ_{\calW}^{(i)},\empP{i})
+W_p(\bbP_i,\empP{i})
\Bigr) \notag\\
\overset{\mathrm{(d)}}{\leq}{}&
2L_\xi^\ell\sum_{i=1}^n
\widehat{\alpha}_i\varepsilon_i
\leq 2L_\xi^\ell r.
\end{align}
Here, (a) follows because $\frac{1}{L_\xi^\ell}\ell(\xdro,\cdot)$
is an admissible $1$-Lipschitz test function; (b) follows from Kantorovich--Rubinstein duality~\cite[Section~11.8]{dudley2018real}; (c) uses $W_1\leq W_p$ for
$p\geq1$~\cite[Remark~6.6]{villani2009optimal}; and (d) follows from $\bbQ_{\calW}^{(i)}\in\calW_i(\varepsilon_i)$
by construction and $\bbP_i\in\calW_i(\varepsilon_i)$ on
$\bigcap_{i=1}^n E_i$.
The inequality preceding (d) is the triangle inequality,
and the final bound uses $\widehat{\alpha}\in\Delta_{n-1}$ and
$r=\max_{i\in[n]}\varepsilon_i$.
For the heterogeneity term in~\eqref{eq:OOS_separation}, since $\widehat\alpha,\alpha^{\star}\in\Delta_{n-1}$, Lemma~\ref{lm:inequality} gives
\begin{align}
&\left|
\sum_{i=1}^n
(\widehat{\alpha}_i-\alpha_i^\star)
\ev_{\bbP_i}[\ell(\xdro,\xi)]
\right|
\notag\\
\leq{}&
\frac{1}{2}
\left(
\max_i\ev_{\bbP_i}[\ell(\xdro,\xi)]
-\min_i\ev_{\bbP_i}[\ell(\xdro,\xi)]
\right)
\|\widehat{\alpha}-\alpha^\star\|_1
\notag\\
\leq{}&
\max_i\ev_{\bbP_i}[\ell(\xdro,\xi)]
-
\min_i\ev_{\bbP_i}[\ell(\xdro,\xi)],
\end{align}
where the last inequality follows from $\|\widehat{\alpha}-\alpha^\star\|_1\leq\|\widehat{\alpha}\|_1+\|\alpha^\star\|_1=2$.
Assumption~\mref{ass:bounded_heterogeneity} ensures  $\sup_{i,j} W_p(\bbP_i,\bbP_j)\le \delta$.
Applying Kantorovich-Rubinstein duality again:
\begin{align}
&\bigl|\ev_{\bbP_i}[\ell(\xdro,\xi)]-\ev_{\bbP_j}[\ell(\xdro,\xi)]\bigr|\notag\\
&\le L_\xi^\ell\,W_p(\bbP_i,\bbP_j)\leq L_\xi^\ell\delta,
\end{align}
which yields
\begin{align}\label{eq:bound2}
\Bigl|\sum_{i=1}^n(\widehat{\alpha}_i-\alpha_i^\star)\ev_{\bbP_i}[\ell(\xdro,\xi)]\Bigr|\leq L_\xi^\ell\delta.
\end{align}
Combining two bounds~\eqref{eq:bound1} and~\eqref{eq:bound2} and noting that they are evaluated on the event $\bigcap_{i=1}^nE_i$ completes the proof.

\section{Proof of Proposition~\mref{prop:local_robustness}}\label{proof:local_robustness}
 
Since \(\bbP_i^\star(x)\) satisfies
\(W_p(\bbP_i^\star(x),\empP{i})=\widehat r_i(x)\),  we have
\[
    \sup_{\bbQ_i:\,W_p(\bbQ_i,\empP{i})\le \widehat r_i(x)}
    \ev_{\bbQ_i}[\ell(x,\xi)]
    \ge
    \ev_{\bbP_i^\star(x)}[\ell(x,\xi)] .
\]
Conversely, for any \(\bbQ_i\) satisfying
\(W_p(\bbQ_i,\empP{i})\le \widehat r_i(x)\), the optimality of \(\bbP_i^\star(x)\) in the penalized problem gives
\[
\ev_{\bbP_i^\star(x)}[\ell(x,\xi)]
-\rho_i \widehat r_i(x)^p
\ge
\ev_{\bbQ_i}[\ell(x,\xi)]
-\rho_i W_p^p(\bbQ_i,\empP{i}) .
\]
Therefore,
\begin{align*}
    &~\ev_{\bbQ_i}[\ell(x,\xi)]
-
\ev_{\bbP_i^\star(x)}[\ell(x,\xi)]\\
\le&~
\rho_i\left(
W_p^p(\bbQ_i,\empP{i})-\widehat r_i(x)^p
\right)\le 0.
\end{align*}
Taking the supremum over all feasible \(\bbQ_i\) yields the reverse inequality, and the proof is complete.

\section{Proof of Theorem~\mref{thm:convergence}}\label{proof:convergence}
We first establish the auxiliary bounds used in the convergence analysis and then combine them to bound the expected primal--dual gap.

\subsection{Auxiliary lemmas}

We start by recording several properties induced by Assumption~\mref{ass:main}.
Denote $ \rho_{\max}:=\max_{i\in[n]}\rho_i$. Since \(\ell\) and \(c\) are continuous on compact domains, we define the following finite constants:
\[
    M_\ell:=\sup_{x\in\calX,\,\xi\in\Xi}|\ell(x,\xi)|,
    \qquad
    M_c:=\sup_{\zeta,\xi\in\Xi}c(\zeta,\xi).
\]

\begin{lemma}[Induced Regularity and Boundedness]\label{lm:induced_bounds}\label{proof:induced_bounds}
Let Assumption~\mref{ass:main} hold. For every \(i\in[n]\), \(x\in\calX\), and \(\zeta\in\Xi\), the inner problem admits a unique maximizer, denoted by $z_i^\star(x,\zeta)\in\operatorname*{arg\,max}_{\xi\in\Xi}\phi_i(x,\zeta,\xi)$. Define the corresponding exact primal oracle by $G_i^x(x;\zeta):=\nabla_x\ell\bigl(x,z_i^\star(x,\zeta)\bigr)$.
Moreover, there exist constants \(L_\xi^\phi,M_x,M_\alpha \in\bbR_{\ge 0}\) such that, for all \(i\in[n]\),
\begin{subequations}
\begin{align}
    &|\phi_i(x,\zeta,\xi)-\phi_i(x,\zeta,\xi')|
    \le L_\xi^\phi\|\xi-\xi'\|,
    \label{eq:phi_lipschitz}\\
    &\nabla F_i(x)
    =\ev_{\zeta\sim\empP{i}}
    \left[G_i^x(x;\zeta)\right],
    \label{eq:unbiased_primal_oracle}\\
    &\|\nabla F_i(x)\|\le M_x,
    \label{eq:bounded_Fi_gradient}\\
    &\|\nabla_\alpha F(x,\alpha)\|\le M_\alpha.
    \label{eq:bounded_dual_gradient}
\end{align}
\end{subequations}
In particular, one may take
\[
    L_\xi^\phi:=L_\xi^\ell+\rho_{\max}L_\xi^c,
    \qquad
    M_x:=M_x^\ell,
\]
and
\[
M_\alpha:=\sqrt n\,(M_\ell+\rho_{\max}M_c).\]
\end{lemma}

For any fixed \(i\in[n]\), \(x\in\calX\), and
\(\zeta\in\Xi\), the mapping $\xi\mapsto\phi_i(x,\zeta,\xi)$ is continuous and strictly concave on the compact convex set \(\Xi\) by Assumption~\mref{ass:main}(i)--(iv). Hence its maximum is obtained at a unique point, denoted by \(z_i^\star(x,\zeta)\). For any \(\xi,\xi'\in\Xi\), the definition of \(\phi_i\) gives
\begin{align*}
    &|\phi_i(x,\zeta,\xi)-\phi_i(x,\zeta,\xi')|\\
    &\le |\ell(x,\xi)-\ell(x,\xi')|
    +\rho_i|c(\zeta,\xi)-c(\zeta,\xi')|\\
    &\le (L_\xi^\ell+\rho_iL_\xi^c)\|\xi-\xi'\|\\
    &\le L_\xi^\phi\|\xi-\xi'\|,
\end{align*}
which proves~\eqref{eq:phi_lipschitz}. By Danskin's theorem~\cite{Bertsekas2016Nonlinear}, the gradient of \(f_i\) with respect to \(x\) is given by $\nabla_x f_i(x,\zeta)=\nabla_x\ell\bigl(x,z_i^\star(x,\zeta)\bigr)$.
Since \(\empP{i}\) has finite support, differentiating its finite empirical
average yields
\[
    \nabla F_i(x)
    =\ev_{\zeta\sim\empP{i}}
    \left[\nabla_x\ell\bigl(x,z_i^\star(x,\zeta)\bigr)\right],
\]
which proves~\eqref{eq:unbiased_primal_oracle}. Assumption~\mref{ass:main}(ii)
then gives
\begin{align*}
    \|\nabla F_i(x)\|
    &\le\ev_{\zeta\sim\empP{i}}
    \left[\left\|\nabla_x\ell\bigl(x,z_i^\star(x,\zeta)\bigr)\right\|\right]\\
    &\le M_x^\ell=M_x,
\end{align*}
proving~\eqref{eq:bounded_Fi_gradient}.
Finally, the definitions of
\(M_\ell\) and \(M_c\) imply
\[
    |f_i(x,\zeta)|=\Big|\sup_{\xi\in\Xi}\{\ell(x,\xi)-\rho_i c(\zeta,\xi)\}\Big|\le M_\ell+\rho_{\max}M_c\,,
\]
which yields $|F_i(x)|\le M_\ell+\rho_{\max}M_c$. Since
\(\nabla_\alpha F(x,\alpha)=(F_1(x),\ldots,F_n(x))^\top\), we have
\[
    \|\nabla_\alpha F(x,\alpha)\|
    \le\sqrt n\,(M_\ell+\rho_{\max}M_c)=M_\alpha,
\]
which proves~\eqref{eq:bounded_dual_gradient}.

\begin{lemma}[Moment Bounds for Exact Stochastic Oracles]\label{lem:exact_oracle_moments}\label{proof:exact_oracle_moments}
Let Assumption~\mref{ass:main} hold. Let
\[
    \sigma_x^2:=(M_x^\ell)^2,
    \quad
    \sigma_\alpha^2:=\frac{n^2}{q}(M_\ell+\rho_{\max}M_c)^2.
\]
Then, for every \(i\in[n]\), we have
\begin{align}
    \ev_{\zeta\sim\empP{i}}
    \left[\|G_i^x(x;\zeta)-\nabla F_i(x)\|^2\right]
    &\le\sigma_x^2.
    \label{eq:var_x}
\end{align}

In addition, let \(\calU\subseteq[n]\), \(|\calU|=q\), be sampled
uniformly from all \(q\)-client subsets of \([n]\). Conditional on
\(\calU\), let every selected client \(i\in\calU\) independently draw
\(\zeta_i\sim\empP{i}\). Recall that \(e_i\in\bbR^n\) denotes the \(i\)th standard basis vector. Define
\[
    V(x):=\frac{n}{q}\sum_{i\in\calU}f_i(x,\zeta_i)e_i.
\]
Then, for every \(x\in\calX\) and \(\alpha\in\Delta_{n-1}\),
\begin{subequations}
    \begin{align}
    &\ev_{\calU,\{\zeta_i\}_{i\in\calU}}[V(x)]
    =\nabla_\alpha F(x,\alpha),
    \label{eq:unbiased_dual_oracle}\\
    &\ev_{\calU,\{\zeta_i\}_{i\in\calU}}
    \left[\|V(x)-\nabla_\alpha F(x,\alpha)\|^2\right]
    \le\sigma_\alpha^2.
    \label{eq:var_alpha}
\end{align}
\end{subequations}
\end{lemma}

Assumption~\mref{ass:main}(ii) implies
\[
    \|G_i^x(x;\zeta)\|
    =\left\|\nabla_x\ell\bigl(x,z_i^\star(x,\zeta)\bigr)\right\|
    \le M_x^\ell.
\]
By~\eqref{eq:unbiased_primal_oracle}, the variance identity therefore yields
\begin{align*}
    &\ev_{\zeta\sim\empP{i}}
    \left[\|G_i^x(x;\zeta)-\nabla F_i(x)\|^2\right]\\
    &=\ev_{\zeta\sim\empP{i}}[\|G_i^x(x;\zeta)\|^2]
    -\|\nabla F_i(x)\|^2\\
    &\le(M_x^\ell)^2=\sigma_x^2,
\end{align*}
which proves~\eqref{eq:var_x}.

For the dual oracle, uniform subset sampling gives
\(\bbP(i\in\calU)=q/n\). Hence, coordinatewise,
\begin{align*}
    \ev[V_i(x)]
    &=\frac{n}{q}\bbP(i\in\calU)
    \ev_{\zeta_i\sim\empP{i}}[f_i(x,\zeta_i)]=F_i(x).
\end{align*}
Consequently, by $\ev[V(x)]=(F_1(x),\ldots,F_n(x))^\top=\nabla_\alpha F(x,\alpha)$, \eqref{eq:unbiased_dual_oracle} is proved.

By Lemma~\ref{lm:induced_bounds},
\(|f_i(x,\zeta_i)|\le M_\ell+\rho_{\max}M_c\).
The coordinate vectors \(e_i\) are mutually orthogonal, and therefore
\begin{align*}
    \|V(x)\|^2
    &=\left(\frac{n}{q}\right)^2
    \sum_{i\in\calU}f_i(x,\zeta_i)^2\\
    &\le\left(\frac{n}{q}\right)^2q(M_\ell+\rho_{\max}M_c)^2\\
    &=\frac{n^2}{q}(M_\ell+\rho_{\max}M_c)^2.
\end{align*}
Using the variance identity once more,
\begin{align*}
    &\ev\left[\|V(x)-\nabla_\alpha F(x,\alpha)\|^2\right]\\
    &=\ev[\|V(x)\|^2]-\|\nabla_\alpha F(x,\alpha)\|^2\\
    &\le\frac{n^2}{q}(M_\ell+\rho_{\max}M_c)^2=\sigma_\alpha^2,
\end{align*}
which proves~\eqref{eq:var_alpha}.

We next introduce the auxiliary quantities used to bound the aggregated and inexact estimators.
Recall that \(\mathcal I_s=\{s\tau,\ldots,(s+1)\tau-1\}\) denotes the iteration block associated with communication round \(s\), and \(s_t\) denotes the unique communication round associated with iteration \(t\).
The client batch sampled at this round is denoted by
\(\calC^{(s_t)}\). Sums over \(i\in\calC^{(s_t)}\) are taken over the
batch entries, with repeated clients counted with multiplicity. We define the
following virtual average model and auxiliary quantities for each iteration \(t\):
    \begin{align}
        x^{(t)}&:= \frac{1}{q}\sum_{i\in\calC^{(s_t)}} x_i^{(t)},
\label{eq:barx_def}\\[2mm]
	\bar{g}_i^{(t)}
	&:= \nabla F_i(x_i^{(t)}),\\[1mm]
	\tilde{g}_i^{(t)}&:=G_i^x(x_i^{(t)};\widehat{\zeta}_i^{(t)}),\\[1mm]
	g_i^{(t)}&:=\nabla_x\ell(x_i^{(t)},z_i^{(t)}),\\[2mm]
	\bar{g}^{(t)}
	&:= \frac{1}{q}\sum_{i\in\calC^{(s_t)}} \bar{g}_i^{(t)},
	\label{eq:barg_def}\\[2mm]
	\tilde g^{(t)}&:=\frac{1}{q}\sum_{i\in\calC^{(s_t)}} \tilde{g}_i^{(t)}, \label{eq:G_def}\\[2mm]
	g^{(t)}
	&:= \frac{1}{q}\sum_{i\in\calC^{(s_t)}} g_i^{(t)},
	\label{eq:gtilde_def}\\[2mm]
	\bar{v}^{(t)}
	&:= \nabla_{\alpha}F(x^{(t)},\alpha^{(s_t)})
= \bigl[F_1(x^{(t)}),\ldots,F_n(x^{(t)})\bigr]^\top,\label{eq:barv_def}\\[2mm]
	\tilde{v}^{(t)}
	&:=\frac{n}{q}\sum_{i\in\calU^{(s_t)}}f_i(x^{(t)},\widehat{\zeta}_i^{(t)})e_i,\label{eq:vtilde_def}\\[2mm]
	v^{(t)}
	    &:=\frac{n}{q}\sum_{i\in\calU^{(s_t)}}\phi_i(x^{(t)},\widehat{\zeta}_i^{(t)},z_i^{(t)})e_i,\label{eq:v_def}\\[2mm]
    \overline{\kappa}^{(s)}
&:= \sum_{t\in\mathcal I_s} \eta_\alpha\, \bar{v}^{(t)},\label{eq:Delta_def}\\[1mm]
\tilde{\kappa}^{(s)}&= \tau\,\eta_\alpha \tilde{v}^{(t^\prime)}\\[1mm]
\kappa^{(s)}&= \tau\,\eta_\alpha v^{(t^\prime)}\\[2mm]
\gamma^{(t)}
&:= \frac{1}{q}\sum_{i\in\calC^{(s_t)}}\bigl\|x_i^{(t)}-x^{(t)}\bigr\|^2,\label{eq:delta_def}
\end{align}
where $\widehat{\zeta}_i^{(t)}$ is uniformly sampled from $i$th local dataset $\calD_i$ at iteration $t$, and, in communication round \(s\), $t^\prime$ is the snapshot index sampled from \(\mathcal I_s\).

    We next present some technical lemmas used later.
\begin{lemma}[Aggregated and Inexact Estimator Bounds]\label{lm:bounds}\label{proof:bounds}
    The exact stochastic primal oracle $\tilde{g}^{(t)}$ and the exact stochastic dual oracle $\tilde v^{(t)}$ are unbiased with bounded variances. Consequently, the snapshot estimator $\tilde{\kappa}^{(s)}$ is unbiased in expectation, while the implemented gradients $g^{(t)}$ and $\kappa^{(s)}$ have bounded variances, which implies
    \begin{subequations}
        {\color{red}
\begin{align}
             &\ev_{\widehat{\zeta}_i^{(t)},\calC^{(s_t)}}[\tilde{g}^{(t)}]=\ev_{\calC^{(s_t)}}[\bar{g}^{(t)}]\notag\\
             &\qquad\qquad=\ev_{\calC^{(s_t)}}\Big[\sum_{i=1}^n\alpha_i^{(s_t)}\nabla F_i(x_i^{(t)})\Big]\,,\label{eq:lm_g_def}\\
             &\ev[\|\tilde g^{(t)}-\bar g^{(t)}\|^2]\leq \frac{\sigma_x^2}{q}\,,\\
             &\ev[\|g^{(t)}-\bar{g}^{(t)}\|^2]\leq 2(L_{\xi}^{g})^2\epsilon^2+\frac{2\sigma_x^2}{q}\,,\\
             &\ev[\tilde v^{(t)}]=\ev[\bar v^{(t)}]\,,\\
             &\ev[\|\tilde v^{(t)}-\bar v^{(t)}\|^2]\leq \sigma_\alpha^2\,,\\
             &\ev[\tilde{\kappa}^{(s)}]=\ev[\bar{\kappa}^{(s)}]\,,\\
		             &\ev[\|\kappa^{(s)}-\bar{\kappa}^{(s)}\|^2]\notag\\
             &\qquad\leq 3\tau^2\eta_\alpha^2\Big(n^2(L_{\xi}^{\phi})^2\epsilon^2+\sigma_\alpha^2+M_\alpha^2\Big)\,.
        \end{align}
}
    \end{subequations}
\end{lemma}
    In~\eqref{eq:lm_g_def}, the first equality follows from
    \eqref{eq:unbiased_primal_oracle} in Lemma~\ref{lm:induced_bounds},
    whereas the second equality follows from the fact that the client batch
    $\calC^{(s_t)}$ is sampled according to $\alpha^{(s_t)}$. Conditional on
    the sampled client batch and current iterates, the local data
    samples are independent, and hence
    \begin{align*}
        &{\ev_{\widehat{\zeta}_i^{(t)}}[\tilde g^{(t)}\mid
        \calC^{(s_t)},\{x_i^{(t)}\}_{i\in\calC^{(s_t)}}]=\bar g^{(t)},}\\
        &{\ev_{\widehat{\zeta}_i^{(t)}}[\|\tilde g^{(t)}-\bar g^{(t)}\|^2\mid
        \calC^{(s_t)},\{x_i^{(t)}\}_{i\in\calC^{(s_t)}}]\leq\frac{\sigma_x^2}{q}.}
    \end{align*}
    Taking total expectation gives the first two primal-oracle claims.
    The variance of $g^{(t)}$ can be bounded by
    \begin{align}
        &\ev[\|g^{(t)}-\bar{g}^{(t)}\|^2]\notag\\
        &= \ev[\|g^{(t)}-\tilde{g}^{(t)}+\tilde{g}^{(t)}-\bar{g}^{(t)}\|^2]\notag\\
        &\leq \Big[\ev\big(\|\frac{1}{q}\sum_{i\in\calC^{(s_t)}}\nabla_x \ell(x_i^{(t)},z_i^{(t)})-\nabla_x \ell(x_i^{(t)},z_i^{\star,(t)})\|^2\big)^{\frac{1}{2}}\notag\\
        &+\ev\big(\|\frac{1}{q}\sum_{i\in\calC^{(s_t)}}(\nabla_x \ell(x_i^{(t)},z_i^{\star,(t)})-\nabla F_i(x_i^{(t)}))\|^2\big)^{\frac{1}{2}}\Big]^2\notag\\
        &\leq (L_{\xi}^{g}\epsilon+\frac{1}{\sqrt{q}}\sigma_x)^2\notag\\
        &\le 2(L_{\xi}^{g})^2\epsilon^2+\frac{2\sigma_x^2}{q}
    \end{align}
where the first inequality follows from Cauchy--Schwarz inequality and the
triangle inequality. The second inequality follows from the Lipschitz
continuity of $\nabla_x\ell(x,\xi)$ over \(\xi\) and the primal-oracle moment
bound~\eqref{eq:var_x} in Lemma~\ref{lem:exact_oracle_moments}.

    For the exact stochastic dual oracle, taking conditional expectation over $\calU^{(s_t)}$ gives
    \begin{align}
        &\ev_{\calU^{(s_t)}}[\tilde v^{(t)}\mid x^{(t)},\{\widehat\zeta_i^{(t)}\}_{i=1}^n]\notag\\
        =&~\ev_{\calU^{(s_t)}}\Big[\frac nq\sum_{i\in\calU^{(s_t)}}f_i(x^{(t)},\widehat\zeta_i^{(t)})e_i\mid x^{(t)},\{\widehat\zeta_i^{(t)}\}_{i=1}^n\Big]\notag\\
        =&~\frac nq\sum_{i=1}^n\bbP(i\in\calU^{(s_t)})f_i(x^{(t)},\widehat\zeta_i^{(t)})e_i\notag\\
        =&~\frac nq\sum_{i=1}^n\frac qn f_i(x^{(t)},\widehat\zeta_i^{(t)})e_i
\notag\\
=&~\bigl[f_1(x^{(t)},\widehat\zeta_1^{(t)}),\ldots,
        f_n(x^{(t)},\widehat\zeta_n^{(t)})\bigr]^\top,
    \end{align}
    The third equality follows since \(\calU^{(s_t)}\) is sampled uniformly
    from all \(q\)-client subsets of \([n]\).

    For each communication round \(s\), let
\[
\mathscr G_s:=\sigma\!\left(
\alpha^{(s)},\{x^{(t)}\}_{t\in\mathcal I_s}
\right)
\]
denote the information generated by the current mixture parameter and the
primal trajectory. Since the client subset and local data samples used by the
    dual estimator are independent of the primal trajectory, taking their
    conditional expectation further yields
    \begin{align}\label{eq:unbiased_dual_gradient}
        \ev[\tilde v^{(t)}\mid\mathscr G_{s_t}]
        &=
        \bigl[F_1(x^{(t)}),\ldots,F_n(x^{(t)})\bigr]^\top
        =\bar v^{(t)}.
    \end{align}
    Taking expectation once more yields
    $\ev[\tilde v^{(t)}]=\ev[\bar v^{(t)}]$.
    The variance bound
    $\ev[\|\tilde v^{(t)}-\bar v^{(t)}\|^2]\leq\sigma_\alpha^2$
    follows by applying~\eqref{eq:var_alpha} in
    Lemma~\ref{lem:exact_oracle_moments} conditionally on \(x^{(t)}\).
By the conditional unbiasedness established in~\eqref{eq:unbiased_dual_gradient} and the uniform snapshot sampling, the snapshot estimator is
unbiased for the accumulated exact dual update:
\begin{equation}\label{eq:conditional_snapshot_unbiasedness}
\begin{aligned}
\ev[\tilde\kappa^{(s)}\mid\mathscr G_s]
&=
\tau\eta_\alpha
\ev[\tilde v^{(t')}\mid\mathscr G_s]\\
&=
\tau\eta_\alpha
\sum_{t\in\mathcal I_s}
\bbP(t'=t\mid\mathscr G_s)
\ev[
\tilde v^{(t')}
\mid\mathscr G_s,t'=t]\\
&=
\tau\eta_\alpha
\sum_{t\in\mathcal I_s}
\frac1\tau
\ev[
\tilde v^{(t)}
\mid\mathscr G_s]\\
&=
\tau\eta_\alpha
\sum_{t\in\mathcal I_s}
\frac1\tau
\bar v^{(t)}=
\bar\kappa^{(s)}.
\end{aligned}
\end{equation}
Taking expectation on both sides yields
$\ev[\tilde\kappa^{(s)}]=\ev[\bar\kappa^{(s)}]$.
	Similarly, the variance of $\kappa^{(s)}$ is decomposed into the inner approximation error, the exact stochastic estimator error, and the random snapshot time-sampling error:
	        \begin{align}
	                &\ev[\|\kappa^{(s)}-\bar{\kappa}^{(s)}\|^2]\notag\\
	                &=\ev\Big[\Big\|\tau\eta_\alpha(v^{(t^\prime)}-\tilde v^{(t^\prime)})
	                +\tau\eta_\alpha(\tilde v^{(t^\prime)}-\bar v^{(t^\prime)})
                    \notag\\
                    &~+\eta_\alpha\Big(\tau\bar v^{(t^\prime)}-\sum_{t\in\mathcal I_s}\bar v^{(t)}\Big)\Big\|^2\Big]\notag\\
	                &\leq 3\tau^2\eta_\alpha^2\ev[\|v^{(t^\prime)}-\tilde v^{(t^\prime)}\|^2]
	                +3\tau^2\eta_\alpha^2\ev[\|\tilde v^{(t^\prime)}-\bar v^{(t^\prime)}\|^2]\notag\\
	                &\quad +3\eta_\alpha^2
	                \ev\Big[\Big\|\tau\bar v^{(t^\prime)}-\sum_{t\in\mathcal I_s}\bar v^{(t)}\Big\|^2\Big]\notag\\
		                &\leq 3\tau^2\eta_\alpha^2\Big(n^2(L_{\xi}^{\phi})^2\epsilon^2+\sigma_\alpha^2+M_\alpha^2\Big)\,.
	        \end{align}
	        The last inequality uses the exact dual-oracle moment
            bound~\eqref{eq:var_alpha} in
            Lemma~\ref{lem:exact_oracle_moments}, the approximation error
            $\|v^{(t^\prime)}-\tilde v^{(t^\prime)}\|
            \le nL_{\xi}^{\phi}\epsilon$, and the bound
            \(\|\bar v^{(t)}\|\le M_\alpha\) from
            Lemma~\ref{lm:induced_bounds}, together with
            \begin{align*}
                &\ev_{t^\prime}\Big[\Big\|\tau\bar v^{(t^\prime)}-\sum_{t\in\mathcal I_s}\bar v^{(t)}\Big\|^2\Big]
	        \\
            =&~
            \tau^2\Big(
            \ev_{t^\prime}\big[\|\bar v^{(t^\prime)}\|^2\big]
            -\Big\|\frac1\tau\sum_{t\in\mathcal I_s}\bar v^{(t)}\Big\|^2
            \Big)
            \\
            \leq&~
            \tau^2\ev_{t^\prime}\big[\|\bar v^{(t^\prime)}\|^2\big]
	        \le \tau^2 M_\alpha^2 .
            \end{align*}
            where the equality follows from the variance identity since $t^\prime$ is sampled uniformly.

The preceding lemmas provide the regularity and estimator bounds needed below. We next establish a one-step primal bound, a local-drift bound, and a one-round dual bound.
For compactness, given the weighted gradient dissimilarity $\Gamma$ in~\meqref{eq:Gamma_def}, define the problem-dependent constants
\[
\begin{aligned}
M_{\mathrm p}&:=M_x^2+\sigma_x^2+(L_{\xi}^{g})^2\epsilon^2,\\
M_{\mathrm d}&:=3(L_{\xi}^{g})^2\epsilon^2+\sigma_x^2+\frac{2\sigma_x^2}{q}+\Gamma.
\end{aligned}
\]

\begin{lemma}[One-Step Primal Bound]\label{lm:primal_step}\label{proof:primal_step}
    For Algorithm~\mref{alg:fedavg_DRO}, under the same conditions as in Theorem~\mref{thm:convergence}, for $\forall x\in\calX$, the following holds:
        \begin{align}\label{eq:lemma_primal_iter}
                &\ev[\|x^{(t+1)}-x\|^2]\notag\\
                \leq&~ \ev[\|x^{(t)}-x\|^2]+(2+L_{x}^{G}\eta_x)\ev[\gamma^{(t)}]\notag\\
                &~-2\eta_x\ev\Big[F(x^{(t)},\alpha^{(s_t)})-F(x,\alpha^{(s_t)})\Big]\notag\\
                &~+2\eta_xD_{\calX}L_{\xi}^{g}\epsilon+6\eta_x^2M_{\mathrm p}
                +4\eta_x^2\tau M_x\sqrt{3M_{\mathrm p}}\,.
        \end{align}
    \end{lemma}
     
    From the updating rule~\meqref{eq:local_update}, we have
        \begin{align}\label{eq:primal_update_analysis}
                &~\ev[\|x^{(t+1)}-x\|^2]\notag\\
                =&~\ev[\|\frac{1}{q}\sum_{i\in\calC^{(s_t)}}\mathrm{Proj}_{\calX}[x_i^{(t)}-\eta_xg_i^{(t)}]-x\|^2]\notag\\
                \overset{{\mathrm{(a)}}}{\leq}&~\ev[\frac{1}{q}\sum_{i\in\calC^{(s_t)}}\|\mathrm{Proj}_{\calX}[x_i^{(t)}-\eta_xg_i^{(t)}]-x\|^2]\notag\\
                \overset{{\mathrm{(b)}}}{\leq}&~\ev[\frac{1}{q}\sum_{i\in\calC^{(s_t)}}\|x_i^{(t)}-\eta_xg_i^{(t)}-x\|^2]\notag\\
                =&~\ev[\frac{1}{q}\sum_{i\in\calC^{(s_t)}}\|x^{(t)}-x+x_i^{(t)}-x^{(t)}-\eta_xg_i^{(t)}\|^2]\notag\\
                =&~\ev[\|x^{(t)}-x\|^2]+\ev[\gamma^{(t)}]+\eta_x^2\ev[\frac{1}{q}\sum_{i\in\calC^{(s_t)}}\|g_i^{(t)}\|^2]\notag\\
                &~{+2\ev\Big[\Big\langle x^{(t)}-x,
                \frac{1}{q}\sum_{i\in\calC^{(s_t)}}
                \big(x_i^{(t)}-x^{(t)}\big)\Big\rangle\Big]}\notag\\
                &~-2\eta_x\ev[\frac{1}{q}\sum_{i\in\calC^{(s_t)}}
                \langle x_i^{(t)}-x^{(t)},g_i^{(t)}\rangle]\notag\\
                &~-2\eta_x\ev[\langle x^{(t)}-x,g^{(t)}\rangle]\notag\\
                \overset{{\mathrm{(c)}}}{=}&~{\ev[\|x^{(t)}-x\|^2]+\ev[\gamma^{(t)}]+\eta_x^2\ev[\frac{1}{q}\sum_{i\in\calC^{(s_t)}}\|g_i^{(t)}\|^2]}\notag\\
                &~{-2\eta_x\ev[\frac{1}{q}\sum_{i\in\calC^{(s_t)}}
                \langle x_i^{(t)}-x^{(t)},g_i^{(t)}\rangle]}\notag\\
                &~{-2\eta_x\ev[\langle x^{(t)}-x,g^{(t)}\rangle]}\notag\\
                \overset{{\mathrm{(d)}}}{\leq}&~\ev[\|x^{(t)}-x\|^2]+\ev[\gamma^{(t)}]+\eta_x^2\ev[\frac{1}{q}\sum_{i\in\calC^{(s_t)}}\|g_i^{(t)}\|^2]\notag\\
                &~+\ev[\frac{1}{q}\sum_{i\in\calC^{(s_t)}}\Big(\|x_i^{(t)}-x^{(t)}\|^2+\eta_x^2\|g_i^{(t)}\|^2\Big)]\notag\\
                &~-2\eta_x\ev[\langle x^{(t)}-x,g^{(t)}\rangle]\notag\\
                =&~\ev[\|x^{(t)}-x\|^2]+2\ev[\gamma^{(t)}]
                \underbrace{-2\eta_x\ev[\langle x^{(t)}-x,g^{(t)}\rangle]}_{T_1}\notag\\
                &~+2\eta_x^2\underbrace{\ev[\frac{1}{q}\sum_{i\in\calC^{(s_t)}}\|g_i^{(t)}\|^2]}_{T_2}\,,
        \end{align}
        where (a) follows from Jensen's inequality; (b) uses the
        nonexpansiveness of the projection; (c) uses the
        identity $q^{-1}\sum_{i\in\calC^{(s_t)}}
        (x_i^{(t)}-x^{(t)})=0$, which follows from~\eqref{eq:barx_def};
        and (d) applies Young's inequality.
        We denote $e_g^{(t)}:=g^{(t)}-\tilde{g}^{(t)}$ as the gap induced by the approximation of the innermost maximization problem, yielding a bound:
        \begin{align}
	                \|e_g^{(t)}\|&=\Big\|\frac{1}{q}\sum_{i\in\calC^{(s_t)}}(g_i^{(t)}-\tilde{g}_i^{(t)})\Big\|\notag\\
	                &\leq\frac{1}{q}\sum_{i\in\calC^{(s_t)}}\Big\|g_i^{(t)}-\tilde{g}_i^{(t)}\Big\|\notag\\
                &\leq L_{\xi}^{g}\epsilon\,,
        \end{align}
        where the last inequality follows due to the Lipschitz continuity of $\xi\mapsto\nabla_x\ell(x,\xi)$.
        
        We first establish a bound for $T_2$.
        For each selected client, denote
        \[
                \delta_i^{z,(t)}:=g_i^{(t)}-\tilde{g}_i^{(t)},
                \qquad
	                \delta_i^{\xi,(t)}:=\tilde{g}_i^{(t)}-\bar{g}_i^{(t)}.
        \]
        We then move to bound $T_2$ as follows:
        \begin{align}
                T_2
                &=\ev\Big[\frac{1}{q}\sum_{i\in\calC^{(s_t)}}\|g_i^{(t)}\|^2\Big]\notag\\
                &=\ev\Big[\frac{1}{q}\sum_{i\in\calC^{(s_t)}}
	                \|\delta_i^{z,(t)}+\delta_i^{\xi,(t)}+\bar{g}_i^{(t)}\|^2\Big]\notag\\
                &\leq3\ev\biggl[\frac{1}{q}\sum_{i\in\calC^{(s_t)}}
                \Bigl(\|\delta_i^{z,(t)}\|^2+\|\delta_i^{\xi,(t)}\|^2+\|\bar{g}_i^{(t)}\|^2\Bigr)\biggr]\notag\\
                &\leq 3M_{\mathrm p}\,,
        \end{align}
        where the three terms correspond to the inner maximization error, the stochastic-sampling error, and the norm of the local gradient, respectively. They are bounded by $(L_{\xi}^{g})^2\epsilon^2$, $\sigma_x^2$, and $M_x^2$, and hence by $M_{\mathrm p}$.

        Now, we are ready to bound $T_1$:
\begin{align}
                T_1=&~-2\eta_x\ev[\langle x^{(t)}-x,\bar{g}^{(t)}+(\tilde{g}^{(t)}-\bar{g}^{(t)})+e_g^{(t)}\rangle]\notag\\
                \overset{\mathrm{(a)}}{=}&~-2\eta_x\ev[\langle x^{(t)}-x,\bar{g}^{(t)}\rangle]
                -2\eta_x\ev[\langle x^{(t)}-x,e_g^{(t)}\rangle]\notag\\
		                \leq&~ -2\eta_x\ev[\langle x^{(t)}-x,\frac{1}{q}\sum_{i\in\calC^{(s_t)}}\bar{g}_i^{(t)}\rangle]\notag\\
                &~+2\eta_x\ev[\|x^{(t)}-x\|\|e_g^{(t)}\|]\notag\\
                \leq&~-2\eta_x\ev[\frac{1}{q}\sum_{i\in\calC^{(s_t)}}\langle x^{(t)}-x_i^{(t)}+x_i^{(t)}-x,\bar{g}_i^{(t)}\rangle]\notag\\
                &~+2\eta_xD_{\calX}L_{\xi}^{g}\epsilon\notag\\
                \overset{\mathrm{(b)}}{\leq}&~-2\eta_x\ev\Big[\frac{1}{q}\sum_{i\in\calC^{(s_t)}}\Big(
                F_i(x^{(t)})-F_i(x_i^{(t)})\notag\\
                &\quad-\frac{L_x^G}{2}\|x^{(t)}-x_i^{(t)}\|^2+F_i(x_i^{(t)})-F_i(x)\Big)\Big]
                \notag\\
                &\quad+2\eta_xD_{\calX}L_{\xi}^{g}\epsilon\notag\\
                =&~-2\eta_x\ev\Big[\frac{1}{q}\sum_{i\in\calC^{(s_t)}}\big(F_i(x^{(t)})-F_i(x)\big)\Big]\notag\\
                &~+\eta_xL_{x}^{G}\ev[\gamma^{(t)}]+2\eta_xD_{\calX}L_{\xi}^{g}\epsilon\notag\\
                =&~-2\eta_x\ev\Big[\frac{1}{q}\sum_{i\in\calC^{(s_t)}}F_i(x^{(t)})-F(x^{(t)},\alpha^{(s_t)})\notag\\
                &~+F(x^{(t)},\alpha^{(s_t)})-\frac{1}{q}\sum_{i\in\calC^{(s_t)}}F_i(x)\mathclose{{\Big]}}\notag\\
                &~+\eta_xL_{x}^{G}\ev[\gamma^{(t)}]+2\eta_xD_{\calX}L_{\xi}^{g}\epsilon\notag\\
                =&~-2\eta_x
                \ev[F(x^{(t)},\alpha^{(s_t)})-F(x,\alpha^{(s_t)})]\notag\\
                &~-2\eta_x
                \underbrace{\ev\Big[\frac{1}{q}\sum_{i\in\calC^{(s_t)}}F_i(x^{(t)})
                -F(x^{(t)},\alpha^{(s_t)})\Big]}_{=:R^{(t)}}\notag\\
                &~+\eta_xL_{x}^{G}\ev[\gamma^{(t)}]+2\eta_xD_{\calX}L_{\xi}^{g}\epsilon\,.
                        \end{align}
        For (a), the conditional unbiasedness established in the proof of
        Lemma~\ref{lm:bounds} and the tower property give
\begin{align*}
        &\ev\big[\langle x^{(t)}-x,
        \tilde g^{(t)}-\bar g^{(t)}\rangle\big]\\
        &=\ev\Big[\Big\langle x^{(t)}-x,
        \ev[\tilde g^{(t)}-\bar g^{(t)}\mid
        \calC^{(s_t)},\{x_i^{(t)}\}_{i\in\calC^{(s_t)}}]\Big\rangle\Big]=0,
        \end{align*}
        since $x^{(t)}-x$ is fixed under the same conditioning.
        Inequality (b) follows from the \(L_x^G\)-Lipschitz continuity of \(\nabla F_i\) applied between \(x_i^{(t)}\) and \(x^{(t)}\), together with the convexity applied between \(x_i^{(t)}\) and \(x\).
        The last equality uses
        $\ev[q^{-1}\sum_{i\in\calC^{(s_t)}}F_i(x)]
        =\ev[F(x,\alpha^{(s_t)})]$, since the client batch is sampled
        according to $\alpha^{(s_t)}$.
        Since $\bar{x}^{(s_t)}$ is fixed before sampling $\calC^{(s_t)}$, we have
        \[
        \ev\Big[\frac{1}{q}\sum_{i\in\calC^{(s_t)}}F_i(\bar{x}^{(s_t)})
        -F(\bar{x}^{(s_t)},\alpha^{(s_t)})\Big]=0.
        \]
        Therefore,
        \begin{align}
            |R^{(t)}|
            \leq&~\ev\Big[\frac{1}{q}\sum_{i\in\calC^{(s_t)}}
            \left|F_i(x^{(t)})-F_i(\bar{x}^{(s_t)})\right|
            \notag\\
            &~+\left|F(x^{(t)},\alpha^{(s_t)})-F(\bar{x}^{(s_t)},\alpha^{(s_t)})\right|\Big]\notag\\
            \leq&~2M_x\ev[\|x^{(t)}-\bar{x}^{(s_t)}\|]\notag\\
            \leq&~2M_x\ev\Big[\frac{1}{q}\sum_{i\in\calC^{(s_t)}}
            \|x_i^{(t)}-\bar{x}^{(s_t)}\|\Big]\notag\\
            \leq&~2M_x\eta_x\sum_{k=s_t\tau}^{t-1}
            \ev\Big[\frac{1}{q}\sum_{i\in\calC^{(s_t)}}\|g_i^{(k)}\|\Big]\notag\\
            \leq&~2M_x\eta_x\sum_{k=s_t\tau}^{t-1}
            \Big(\ev\Big[\frac{1}{q}\sum_{i\in\calC^{(s_t)}}\|g_i^{(k)}\|^2\Big]\Big)^{1/2}\notag\\
            \leq&~{2M_x\eta_x\tau\sqrt{3M_{\mathrm p}}}.
        \end{align}
        Here, the second inequality follows from Lemma~\ref{lm:induced_bounds}, which implies $M_x$-Lipschitz continuity of $F_i$ and $F(\cdot,\alpha^{(s_t)})$ on $\calX$. The fourth inequality follows from projection nonexpansiveness and the initialization $x_i^{(s_t\tau)}=\bar{x}^{(s_t)}$ at the beginning of the communication round. The last inequality follows by applying the preceding second-moment bound at every $k\in\mathcal I_{s_t}$ and using $t-s_t\tau\leq\tau$.
        As a result, we have
        \begin{align}
            T_1
            \leq&~
            -2\eta_x\ev\Big[F(x^{(t)},\alpha^{(s_t)})-F(x,\alpha^{(s_t)})\Big]
            \notag\\
            &~+\eta_x L_{x}^{G}\ev[\gamma^{(t)}]+2\eta_xD_{\calX}L_{\xi}^{g}\epsilon
            +4\eta_x^2\tau M_x\sqrt{3M_{\mathrm p}}.
        \end{align}
        By substituting $T_1, T_2$  into~\eqref{eq:primal_update_analysis}, we obtain~\eqref{eq:lemma_primal_iter}, which completes the proof.
     
    The following lemma bounds the deviation between the local models $x_i^{(t)}$ and the virtual global model $x^{(t)}$.

\begin{lemma}[Local Model Drift]\label{lm:local_drift}\label{proof:local_drift}
    For Algorithm~\mref{alg:fedavg_DRO}, suppose that
    $12\eta_x^2\tau^2(L_x^G)^2\leq\frac12$. Then the expected average squared
    distance between the local models and the virtual global model satisfies
    \begin{equation}\label{eq:local_drift_simplified}
        \frac{1}{T}\sum_{t=0}^{T-1}\ev[\gamma^{(t)}]
        \leq 12\eta_x^2\tau^2M_{\mathrm d}.
    \end{equation}
    \end{lemma}
     
        We first recall the finite-sample variance identity. For any vectors
        $z_1,\ldots,z_q$ and their average
        $\bar z:=q^{-1}\sum_{i=1}^qz_i$,
        \begin{align}
        \frac{1}{q}\sum_{i=1}^q\|z_i-\bar z\|^2
        &=\frac{1}{q}\sum_{i=1}^q\|z_i\|^2-\|\bar z\|^2\notag\\
        &=\frac{1}{2q^2}\left(2q\sum_{i=1}^q\|z_i\|^2
        -2\left\|\sum_{i=1}^qz_i\right\|^2\right)\notag\\
        &=\frac{1}{2q^2}\sum_{i,j=1}^q\|z_i-z_j\|^2.
        \label{eq:finite_sample_variance}
        \end{align}
        At the beginning of round $s$, the common initialization gives
        $\gamma^{(s\tau)}=0$. For any
        $t\in\mathcal I_s\setminus\{s\tau\}$, the updating rules yield
\begin{align}
                &~\ev[\gamma^{(t)}]
                =\ev\Big[\frac{1}{q}\sum_{i\in\calC^{(s_t)}}\|x_i^{(t)}-x^{(t)}\|^2\Big]\nonumber\\
                &\overset{\text{(a)}}{=}\ev\Big[\frac{1}{2q^2}\sum_{i,j\in\calC^{(s_t)}}\|x_i^{(t)}-x_j^{(t)}\|^2\Big]\nonumber\\
                &\overset{\text{(b)}}{=}\ev\Big[\frac{1}{2q^2}\sum_{i,j\in\calC^{(s_t)}}\|\mathrm{Proj}_{\calX}(x_i^{(t-1)}-\eta_xg_i^{(t-1)})\notag\\
                &\quad-\mathrm{Proj}_{\calX}(x_j^{(t-1)}-\eta_xg_j^{(t-1)})\|^2\Big]\nonumber\\
                &\overset{\text{(c)}}{\leq}\ev\Big[\frac{1}{2q^2}\sum_{i,j\in\calC^{(s_t)}}
                \Big(\eta_x\sum_{k=s\tau}^{t-1}\|g_i^{(k)}-g_j^{(k)}\|\Big)^2\Big]\nonumber\\
                &\overset{\text{(d)}}{\leq}\ev\Big[\frac{\eta_x^2\tau}{2q^2}\sum_{i,j\in\calC^{(s_t)}}
                \sum_{k\in\mathcal I_s}\|g_i^{(k)}-g_j^{(k)}\|^2\Big]\nonumber\\
	                &\overset{\text{(e)}}{=} \ev\Big[\frac{1}{q}\sum_{i\in\calC^{(s_t)}}\eta_x^2\tau\sum_{k\in\mathcal I_s}\Big\|g_i^{(k)}-\frac{1}{q}\sum_{j\in\calC^{(s_t)}}g_j^{(k)}\Big\|^2\Big]\nonumber\\
	                &=\ev\Big[\frac{1}{q}\sum_{i\in\calC^{(s_t)}}\eta_x^2\tau\sum_{k\in\mathcal I_s}\Big\|(g_i^{(k)}-\tilde g_i^{(k)})\notag\\
&\quad+(\tilde g_i^{(k)}-\bar g_i^{(k)})
+(\bar g_i^{(k)}-\nabla F_i(x^{(k)}))\notag\\
&\quad+\Big(\nabla F_i(x^{(k)})
-\frac{1}{q}\sum_{j\in\calC^{(s_t)}}\nabla F_j(x^{(k)})\Big)\notag\\
&\quad+\Big(\frac{1}{q}\sum_{j\in\calC^{(s_t)}}\nabla F_j(x^{(k)})-\bar g^{(k)}\Big)+(\bar g^{(k)}-g^{(k)})\Big\|^2\Big]\nonumber\\
	                &\overset{\text{(f)}}{\leq} \ev\Big[6\eta_x^2\tau\sum_{k\in\mathcal I_s}\Big[\frac{1}{q}\sum_{i\in\calC^{(s_t)}}\Big\|g_i^{(k)}-\tilde{g}_i^{(k)}\Big\|^2\nonumber\\
                &\quad+\frac{1}{q}\sum_{i\in\calC^{(s_t)}}\Big\|\tilde{g}_i^{(k)}-\bar{g}_i^{(k)}\Big\|^2+\frac{1}{q}\sum_{i\in\calC^{(s_t)}}\Big\|\bar{g}_i^{(k)}-\nabla F_i(x^{(k)})\Big\|^2\nonumber\\
	                &\quad+\frac{1}{q}\sum_{i\in\calC^{(s_t)}}\Big\|\nabla F_i(x^{(k)})-\frac{1}{q}\sum_{j\in\calC^{(s_t)}}\nabla F_j(x^{(k)})\Big\|^2\nonumber\\
                &\quad+\Big\|\frac{1}{q}\sum_{j\in\calC^{(s_t)}}\nabla F_j(x^{(k)})-\bar{g}^{(k)}\Big\|^2+\Big\|\bar{g}^{(k)}-g^{(k)}\Big\|^2\Big]\Big]\,,
                \label{eq:local_drift_six_terms}
            \end{align}
        where equality~(a) applies~\eqref{eq:finite_sample_variance} to the local models with $\bar z=x^{(t)}$. Equality~(b) substitutes the local update rule~\meqref{eq:local_update}. Inequality~(c) recursively applies projection nonexpansiveness and the triangle inequality back to the common initialization $x_i^{(s\tau)}=x_j^{(s\tau)}=\bar{x}^{(s)}$. Inequality~(d) follows from Cauchy--Schwarz, $t-s\tau\leq\tau$, and enlargement of the summation range to $\mathcal I_s$. Equality~(e) applies~\eqref{eq:finite_sample_variance} in the reverse direction to the client gradients at each $k\in\mathcal I_s$. Finally, inequality~(f) uses $\|\sum_{r=1}^{6}a_r\|^2\leq6\sum_{r=1}^{6}\|a_r\|^2$.
        The six terms on the right-hand side of
        \eqref{eq:local_drift_six_terms} are bounded, in their order of
        appearance, using Assumption~\mref{ass:main}(ii) and the
        $\epsilon$-accurate solution, the variance bound~\eqref{eq:var_x},
        Assumption~\mref{ass:main}(v), the definition of $\Gamma$ in~\meqref{eq:Gamma_def},
        Assumption~\mref{ass:main}(v) together with Jensen's inequality, and
        Lemma~\ref{lm:bounds}, respectively. Substituting these bounds gives
        \begin{align}
        \ev[\gamma^{(t)}]
        \leq&~6\eta_x^2\tau\sum_{k\in\mathcal I_s}\Big(
        \underbrace{(L_{\xi}^{g})^2\epsilon^2}_{g_i^{(k)}-\tilde g_i^{(k)}}
        +\underbrace{\sigma_x^2}_{\tilde g_i^{(k)}-\bar g_i^{(k)}}
        \notag\\
        &~+\underbrace{(L_x^G)^2\ev[\gamma^{(k)}]}
        _{\bar g_i^{(k)}-\nabla F_i(x^{(k)})}
        +\underbrace{\Gamma}_{\text{client heterogeneity}}
        \notag\\
        &~+\underbrace{(L_x^G)^2\ev[\gamma^{(k)}]}
        _{q^{-1}\sum_j\nabla F_j(x^{(k)})-\bar g^{(k)}}
        +\underbrace{2(L_{\xi}^{g})^2\epsilon^2
        +\frac{2\sigma_x^2}{q}}_{\bar g^{(k)}-g^{(k)}}\Big)
        \notag\\
        =&~6\eta_x^2\tau\sum_{k\in\mathcal I_s}\Big(
        3(L_{\xi}^{g})^2\epsilon^2+\sigma_x^2
        +\frac{2\sigma_x^2}{q}\notag\\
        &\hspace{31mm}+2(L_x^G)^2\ev[\gamma^{(k)}]+\Gamma\Big).
        \end{align}
        Summing the preceding bound over \(t\in\mathcal I_s\) gives
        \begin{equation}
            \begin{aligned}
            &\sum_{t\in\mathcal I_s} \ev[\gamma^{(t)}]\notag\\
            &\leq 6\eta_x^2\tau\sum_{t\in\mathcal I_s}\sum_{k\in\mathcal I_s}
	            \Big(3(L_{\xi}^{g})^2\epsilon^2+\sigma_x^2+\frac{2\sigma_x^2}{q}\notag\\
	            &\hspace{30mm}+2(L_{x}^{G})^2\ev[\gamma^{(k)}]+\Gamma \Big)\\
            &\leq 6\eta_x^2\tau^2\sum_{k\in\mathcal I_s}
	            \Big(3(L_{\xi}^{g})^2\epsilon^2+\sigma_x^2+\frac{2\sigma_x^2}{q}\\
	            &\hspace{30mm}+2(L_{x}^{G})^2\ev[\gamma^{(k)}]+\Gamma \Big).
            \end{aligned}
        \end{equation}
        Rearranging the preceding inequality and using
        \[
        \textstyle 12\eta_x^2\tau^2(L_x^G)^2\leq\frac12
        \]
        give
        \begin{equation}
            \sum_{t\in\mathcal I_s}\ev[\gamma^{(t)}]
            \leq 12\eta_x^2\tau^2
            \sum_{k\in\mathcal I_s}M_{\mathrm d}.
        \end{equation}
        Summing over $s=0,\ldots,S-1$ and dividing by $T$
        yields~\eqref{eq:local_drift_simplified}.
     
    This lemma bounds the local consensus error among the selected client models over each communication round. We next turn to the dual update.

\begin{samepage}
\begin{lemma}[One-Round Dual Bound]\label{lm:dual_step}\label{proof:dual_step} For Algorithm~\mref{alg:fedavg_DRO}, under the same conditions as in Theorem~\mref{thm:convergence}, for $\forall \alpha\in\Delta_{n-1}$, the following holds:
\begin{align}
	                &\ev[\|\alpha^{(s+1)}-\alpha\|^2]\leq \ev[\|\alpha^{(s)}-\alpha\|^2]+\ev[\|\bar{\kappa}^{(s)}\|^2]\notag\\
                &~+(2\sqrt{2}+2\tau\eta_\alpha M_\alpha)\frac{\tau\eta_\alpha n}{\sqrt{q}}L_{\xi}^{\phi}\epsilon+\ev[\|\kappa^{(s)}-\bar{\kappa}^{(s)}\|^2]\notag\\
                &~-2\eta_\alpha\sum_{t\in\mathcal I_s}\ev[F(x^{(t)},\alpha)-F(x^{(t)},\alpha^{(s)})]
        \end{align}
    \end{lemma}
     
\end{samepage}
    According to the updating rule of $\alpha$ as in~\meqref{eq:update_alpha}, we have
\begin{align}\label{eq:duak_analysis}
    &~\ev[\|\alpha^{(s+1)}-\alpha\|^2]\notag\\
    =&~\ev\Big[\Big\|\mathrm{Proj}_{\Delta_{n-1}}(\alpha^{(s)}+\kappa^{(s)})-\alpha\Big\|^2\Big]\notag\\
    {\overset{\text{(a)}}{\leq}}&~\ev\Big[\|\alpha^{(s)}-\alpha+\kappa^{(s)}\|^2\Big]\notag\\
    =&~\ev\Big[\|\alpha^{(s)}+\bar{\kappa}^{(s)}-\alpha+\kappa^{(s)}-\bar{\kappa}^{(s)}\|^2\Big]\notag\\
    =&~\ev[\|\alpha^{(s)}+\bar{\kappa}^{(s)}-\alpha\|^2]
    +\ev[\|\kappa^{(s)}-\bar{\kappa}^{(s)}\|^2]\notag\\
    &~+{2\ev[\langle
    \alpha^{(s)}+\bar\kappa^{(s)}-\alpha,
    \kappa^{(s)}-\bar\kappa^{(s)}\rangle]}\notag\\
    {\overset{\text{(b)}}{=}}&~
    {\ev[\|\alpha^{(s)}+\bar\kappa^{(s)}-\alpha\|^2]
    +\ev[\|\kappa^{(s)}-\bar\kappa^{(s)}\|^2]}\notag\\
    &~+{\underbrace{2\ev[\langle
    \alpha^{(s)}+\bar\kappa^{(s)}-\alpha,
    \kappa^{(s)}-\tilde\kappa^{(s)}\rangle]}_{T_3}}\notag\\
    =&~\ev[\|\alpha^{(s)}-\alpha\|^2]
    +\ev[\|\bar{\kappa}^{(s)}\|^2]
    +2\ev[\langle \alpha^{(s)}-\alpha,\bar{\kappa}^{(s)}\rangle]\notag\\
    &~+\ev[\|\kappa^{(s)}-\bar{\kappa}^{(s)}\|^2]
	    +T_3\notag\\
    {\overset{\text{(c)}}{=}}&~\ev[\|\alpha^{(s)}-\alpha\|^2]
    +\ev[\|\bar{\kappa}^{(s)}\|^2]
    +\ev[\|\kappa^{(s)}-\bar{\kappa}^{(s)}\|^2]\notag\\
    &~+2\eta_\alpha\sum_{t\in\mathcal I_s}
    \ev[\langle \alpha^{(s)}-\alpha,\nabla_\alpha F(x^{(t)},\alpha^{(s)})\rangle]
    +T_3\notag\\
    {\overset{\text{(d)}}{=}}&~\ev[\|\alpha^{(s)}-\alpha\|^2]
    +\ev[\|\bar{\kappa}^{(s)}\|^2]
    +\ev[\|\kappa^{(s)}-\bar{\kappa}^{(s)}\|^2]\notag\\
    &~-2\eta_\alpha\sum_{t\in\mathcal I_s}
    \ev[F(x^{(t)},\alpha)-F(x^{(t)},\alpha^{(s)})]
    +T_3\,,
	\end{align}
	        Step~(a) uses the nonexpansiveness of projection. For~(b),
	        the conditional identity~\eqref{eq:conditional_snapshot_unbiasedness}
	        gives
	        $\ev[\tilde\kappa^{(s)}-\bar\kappa^{(s)}\mid\mathscr G_s]=0$.
	        Since $\alpha^{(s)}+\bar\kappa^{(s)}-\alpha$ is
	        $\mathscr G_s$-measurable, the tower property yields
	        \begin{align}
	        &\ev\!\left[\left\langle
	        \alpha^{(s)}+\bar\kappa^{(s)}-\alpha,
	        \tilde\kappa^{(s)}-\bar\kappa^{(s)}
	        \right\rangle\right]\notag\\
	        &=\ev\!\left[\left\langle
	        \alpha^{(s)}+\bar\kappa^{(s)}-\alpha,
	        \ev[\tilde\kappa^{(s)}-\bar\kappa^{(s)}\mid\mathscr G_s]
	        \right\rangle\right]=0.
	        \label{eq:dual_cross_zero}
	        \end{align}
	        Equality~(c) follows from the definitions of $\bar\kappa^{(s)}$ and $\tilde\kappa^{(s)}$.
	        Equality~(d) uses the linearity of $F(x,\cdot)$.

            We next bound $T_3$. By the definitions
	$\kappa^{(s)}=\tau\eta_\alpha v^{(t^\prime)}$ and
$\tilde{\kappa}^{(s)}=\tau\eta_\alpha\tilde v^{(t^\prime)}$, we have
        \begin{align}
                T_3
    &=2\ev[\langle \alpha^{(s)}+\bar{\kappa}^{(s)}-\alpha,
    \kappa^{(s)}-\tilde{\kappa}^{(s)}\rangle]\notag\\
    &=2\tau\eta_\alpha
    \ev[\langle \alpha^{(s)}+\bar{\kappa}^{(s)}-\alpha,
    v^{(t^\prime)}-\tilde v^{(t^\prime)}\rangle]\notag\\
    &\leq
    2\tau\eta_\alpha
    \ev[(\|\alpha^{(s)}-\alpha\|+\|\bar{\kappa}^{(s)}\|)
    \|v^{(t^\prime)}-\tilde v^{(t^\prime)}\|]\notag\\
    {\overset{\text{(a)}}{\leq}}&
    2\tau\eta_\alpha
    \Big(\sqrt{2}+\tau\eta_\alpha M_\alpha\Big)
    \ev[\|v^{(t^\prime)}-\tilde v^{(t^\prime)}\|]\notag\\
    {\overset{\text{(b)}}{\leq}}&
    2\tau\eta_\alpha
    \Big(\sqrt{2}+\tau\eta_\alpha M_\alpha\Big)
	    \frac{n}{\sqrt q}L_{\xi}^{\phi}\epsilon\notag\\
	    &=
    \Big(2\sqrt{2}+2\tau\eta_\alpha M_\alpha\Big)
	    \frac{\tau\eta_\alpha n}{\sqrt q}L_{\xi}^{\phi}\epsilon,
        \end{align}
        where inequality~(a) uses
        $\|\alpha^{(s)}-\alpha\|\leq\sqrt{2}$ for points in
        $\Delta_{n-1}$ and
        $\|\bar\kappa^{(s)}\|\leq\tau\eta_\alpha M_\alpha$, where the latter
        follows from the definition of $\bar\kappa^{(s)}$ and the bound on
        $\|\bar v^{(t)}\|$ in Lemma~\ref{lm:induced_bounds}.
        Inequality~(b) follows from~\eqref{eq:phi_lipschitz} in Lemma~\ref{lm:induced_bounds} and
        the $\epsilon$-accurate inner solution~$z_i^{(t')}$. In particular,
\begin{align*}
        \|v^{(t')}-\tilde v^{(t')}\|^2
        =&~\frac{n^2}{q^2}\sum_{i\in\calU^{(s)}}
        \big|\phi_i(x^{(t')},\widehat\zeta_i^{(t')},z_i^{(t')})\notag\\
        &~-\phi_i\!\left(
x^{(t')},\widehat\zeta_i^{(t')},z_i^{\star,(t')}
\right)\big|^2\\
        \leq&~\frac{n^2}{q}(L_\xi^\phi)^2\epsilon^2,
        \end{align*}
        which implies
        $\ev[\|v^{(t')}-\tilde v^{(t')}\|]
        \leq nL_\xi^\phi\epsilon/\sqrt q$.
        Substituting the resulting bound on $T_3$
        into~\eqref{eq:duak_analysis} completes the proof.
     
    \subsection{Completion of the proof}
We now combine the auxiliary bounds to prove Theorem~\mref{thm:convergence}.
The primal-dual gap is decomposed into a dual term and a primal term. By the convexity of $F(\cdot,\alpha)$, the linearity of
	    $F(x,\cdot)$, and the definitions of the averaged iterates, we have
\begin{align}\label{eq:separable_performance_gap}
	        \operatorname{Gap}&
	        (\widehat{x}_T,\widehat{\alpha}_T)
	        =\max_{\alpha\in\Delta_{n-1}}
	        \ev[F(\widehat{x}_T,\alpha)]
	        -\min_{x\in\calX}\ev[F(x,\widehat{\alpha}_T)]\nonumber\\
	        \leq~&\max_{\alpha\in\Delta_{n-1}}\frac1T
	        \sum_{t=0}^{T-1}\ev[F(x^{(t)},\alpha)]\nonumber\\
	        &~-\min_{x\in\calX}\frac1T\sum_{t=0}^{T-1}
	        \ev[F(x,\alpha^{(s_t)})]\nonumber\\
	        =~&\frac1T\max_{\alpha\in\Delta_{n-1}}
	        \sum_{t=0}^{T-1}\ev\Big[
	        F(x^{(t)},\alpha)-F(x^{(t)},\alpha^{(s_t)})\Big]\nonumber\\
	        &~+\frac1T\max_{x\in\calX}\sum_{t=0}^{T-1}
	        \ev\Big[F(x^{(t)},\alpha^{(s_t)})
	        -F(x,\alpha^{(s_t)})\Big].
	    \end{align}
	    The inequality follows from Jensen's inequality.
	    Rearranging Lemma~\ref{lm:dual_step}, summing over
	    $s=0,\ldots,S-1$, and dividing by $T$ yield
        \begin{align}
            &~\frac{1}{T}\sum_{s=0}^{S-1}\sum_{t\in\mathcal I_s}
            \ev\big[F(x^{(t)},\alpha)-F(x^{(t)},\alpha^{(s)})\big]\notag\\
            \leq&~
            \frac{1}{2\eta_\alpha T}\sum_{s=0}^{S-1}
            \Big(\ev[\|\alpha^{(s)}-\alpha\|^2]
            -\ev[\|\alpha^{(s+1)}-\alpha\|^2]\Big)\notag\\
            &~+\frac{1}{2\eta_\alpha T}\sum_{s=0}^{S-1}
            \ev[\|\bar\kappa^{(s)}\|^2]
            +\frac{1}{2\eta_\alpha T}\sum_{s=0}^{S-1}
            \ev[\|\kappa^{(s)}-\bar\kappa^{(s)}\|^2]\notag\\
            &~+\frac{S}{T}(2\sqrt{2}+2\tau\eta_\alpha M_\alpha)
            \frac{\tau\eta_\alpha n}{2\eta_\alpha\sqrt q}
            L_\xi^\phi\epsilon\notag\\
            \overset{\text{(a)}}{\leq}&~
            \frac{1}{2\eta_\alpha T}\sum_{s=0}^{S-1}
            \Big(\ev[\|\alpha^{(s)}-\alpha\|^2]
            -\ev[\|\alpha^{(s+1)}-\alpha\|^2]\Big)\notag\\
            &~+\frac{S\tau^2\eta_\alpha}{2T}M_\alpha^2
            +\frac{3S\tau^2\eta_\alpha}{2T}
            \Big(n^2(L_\xi^\phi)^2\epsilon^2+\sigma_\alpha^2+M_\alpha^2\Big)\notag\\
            &~+\frac{S}{T}(2\sqrt{2}+2\tau\eta_\alpha M_\alpha)
            \frac{\tau\eta_\alpha n}{2\eta_\alpha\sqrt q}
            L_\xi^\phi\epsilon\notag\\
            \overset{\text{(b)}}{\leq}&~
            \frac{1}{\eta_\alpha T}
            +\frac{3}{2}\tau\eta_\alpha
            \Big(n^2(L_\xi^\phi)^2\epsilon^2+\sigma_\alpha^2+M_\alpha^2\Big)
            +\frac{\tau\eta_\alpha}{2}M_\alpha^2\notag\\
            &~+\frac{n}{\sqrt q}
            (\sqrt{2}+\tau\eta_\alpha M_\alpha)L_\xi^\phi\epsilon,
        \end{align}
	    where step~(a) uses the variance bound on
	    $\kappa^{(s)}-\bar\kappa^{(s)}$ from Lemma~\ref{lm:bounds} and
	    $\|\bar\kappa^{(s)}\|\leq\tau\eta_\alpha M_\alpha$ from
	    Lemma~\ref{lm:induced_bounds}. Step~(b) uses the telescoping sum,
	    $S\tau=T$, and
	    $\|\alpha^{(0)}-\alpha\|^2\leq2$ for
	    $\alpha^{(0)},\alpha\in\Delta_{n-1}$.
	    Since the resulting upper bound is independent of the fixed comparator
	    $\alpha$, it also bounds the maximum over
	    $\alpha\in\Delta_{n-1}$ in~\eqref{eq:separable_performance_gap}.

    Similarly, substituting Lemma~\ref{lm:bounds} into
    Lemma~\ref{lm:primal_step} and summing from $t=0$ to $T-1$ yield
\begin{align}
            &~\frac{1}{T}\sum_{t=0}^{T-1}\ev(F(x^{(t)},\alpha^{(s_t)})-F(x,\alpha^{(s_t)}))\notag\\
            \leq&~\frac{1}{T}\sum_{t=0}^{T-1}\frac{1}{2\eta_x}(\ev[\|x^{(t)}-x\|^2]-\ev[\|x^{(t+1)}-x\|^2])\notag\\
            &~+3\eta_xM_{\mathrm p}
            +\frac{1}{T}\sum_{t=0}^{T-1}\Big(\frac{1}{\eta_x}+\frac{L_{x}^{G}}{2}\Big)\ev[\gamma^{(t)}]\notag\\
            &~+D_{\calX}L_{\xi}^{g}\epsilon
            +2\eta_x\tau M_x\sqrt{3M_{\mathrm p}}.
        \end{align}
        We now verify the condition of Lemma~\ref{lm:local_drift}. For the parameter choice in
        Theorem~\mref{thm:convergence},
        $\tau=\max\{1,\lfloor T^{1/8}\rfloor\}\leq T^{1/8}$ for $T\geq1$ and
        $\eta_x=(5L_x^G)^{-1}T^{-5/8}$, and hence
\begin{align*}
        12\eta_x^2\tau^2(L_x^G)^2
        &\leq
        12\left(\frac{T^{-5/8}}{5L_x^G}\right)^2
        T^{1/4}(L_x^G)^2\\*
        &=\frac{12}{25}T^{-1}<\frac{1}{2}.
        \end{align*}
        Therefore, Lemma~\ref{lm:local_drift} applies and gives
        \begin{align}
        \frac{1}{T}\sum_{t=0}^{T-1}
        \left(\frac{1}{\eta_x}+\frac{L_x^G}{2}\right)
        \ev[\gamma^{(t)}]
        \leq&~\eta_x\tau^2(12+6\eta_xL_x^G)M_{\mathrm d}.
        \label{eq:theorem_local_drift_application}
        \end{align}
        Applying~\eqref{eq:theorem_local_drift_application} and the
        telescoping sum to the preceding inequality yields
        \begin{align}
            &~\frac{1}{T}\sum_{t=0}^{T-1}\ev(F(x^{(t)},\alpha^{(s_t)})-F(x,\alpha^{(s_t)}))\notag\\
            \leq&~\frac{D_{\calX}^2}{2\eta_xT}+3\eta_xM_{\mathrm p}+D_{\calX}L_{\xi}^{g}\epsilon
            +2\eta_x\tau M_x\sqrt{3M_{\mathrm p}}\notag\\
            &~+\eta_x\tau^2(12+6\eta_x L_{x}^{G})M_{\mathrm d}.
        \end{align}
        Combining the primal and dual
        bounds therefore yields
        \begin{align}
	            &~\operatorname{Gap}
	            (\widehat{x}_T,\widehat{\alpha}_T)\notag\\
           \leq&
\frac{D_{\calX}^2}{2\eta_xT}
+3\eta_x M_{\mathrm p}
	+D_{\calX}L_{\xi}^{g}\epsilon
    +2\eta_x\tau M_x\sqrt{3M_{\mathrm p}}
	 \notag\\
&+\frac{1}{\eta_\alpha T}
		+\frac{3}{2}\tau\eta_\alpha\Big(n^2(L_{\xi}^{\phi})^2\epsilon^2+\sigma_\alpha^2+M_\alpha^2\Big)
	+\frac{\eta_\alpha\tau}{2}M_\alpha^2 \notag\\
	&+\frac{n}{\sqrt q}(\sqrt2+\tau\eta_\alpha M_\alpha)L_{\xi}^{\phi}\epsilon+\eta_x\tau^2(12+6\eta_xL_{x}^{G})M_{\mathrm d}.
    \end{align}
	Substituting $\eta_x=(5L_x^G)^{-1}T^{-5/8}$,
    $\eta_\alpha=T^{-9/16}$, and $\tau\leq T^{1/8}$ into this bound yields
\begin{align}
	        &\operatorname{Gap}
	        (\widehat{x}_T,\widehat{\alpha}_T)
	            \leq\Big(\frac{5L_{x}^{G}D_{\calX}^2}{2}
	            +\frac{12}{5L_{x}^{G}}M_{\mathrm d}\Big)T^{-3/8}\notag\\*
	            &
		            +\Big(1+\frac{3}{2}n^2(L_{\xi}^{\phi})^2\epsilon^2
                +2M_\alpha^2+\frac{3\sigma_\alpha^2}{2}
		            +\frac{nM_\alpha L_{\xi}^{\phi}\epsilon}{\sqrt q}\Big)T^{-7/16}\notag\\
	            &
                +\frac{2M_x\sqrt{3M_{\mathrm p}}}{5L_{x}^{G}}T^{-1/2}+\frac{3}{5L_{x}^{G}}M_{\mathrm p}T^{-5/8}
	            \notag\\
	            &+\frac{6}{25L_{x}^{G}}M_{\mathrm d}T^{-1}
	            +\Big(D_{\calX}L_{\xi}^{g}
	            +\sqrt{\frac{2}{q}}nL_{\xi}^{\phi}\Big)\epsilon\,.
    \end{align}
    Therefore,
    \[
        \operatorname{Gap}
        (\widehat{x}_T,\widehat{\alpha}_T)
        =
        \calO\Big(T^{-3/8}+\epsilon\Big),
    \]
    which is the claimed bound.

\fi

\bibliographystyle{apacite}
\bibliography{mybibfile}

\end{document}